\documentclass{article}

\usepackage{arxiv}

\usepackage[utf8]{inputenc} 
\usepackage[T1]{fontenc}    
\usepackage{hyperref}       
\usepackage{url}            
\usepackage{booktabs}       
\usepackage{amsfonts}       
\usepackage{nicefrac}  
\usepackage{microtype}      
\usepackage{lipsum}		
\usepackage{graphicx}
\usepackage{natbib}
\usepackage{doi}
\usepackage{graphicx} 
\usepackage{parskip}  
\usepackage{setspace} 
\usepackage{refcount} 
\usepackage{upquote}  
 \usepackage{booktabs}
 \usepackage{tabularx,ragged2e}
\usepackage{xcolor}
\usepackage{listings}
\usepackage{courier}
\usepackage{amsfonts}
\usepackage{amsmath}
\usepackage{natbib}
\usepackage{multirow} 
\usepackage{pifont}
\usepackage{xcolor}
\usepackage{caption} 
\newcommand{\cmark}{\textcolor{green}{\ding{51}}} 
\newcommand{\xmark}{\textcolor{red}{\ding{55}}}   

\usepackage[most]{tcolorbox}
\usepackage[T1]{fontenc}
\usepackage{subfiles}
\newtcolorbox{blockquote}{
  colback=gray!10,
  colframe=gray!50,
  left=2mm,
  right=2mm,
  top=1mm,
  bottom=1mm,
  boxrule=0.5pt,
  fontupper=\ttfamily, 
  sharp corners,
  before skip=10pt,
  after skip=10pt
}
\lstdefinelanguage{json}{
    basicstyle=\normalfont\ttfamily,
    numbers=left,
    numberstyle=\scriptsize,
    breaklines=true,
    frame=lines,
    backgroundcolor=\color{gray!10},
    showstringspaces=false,
    string=[db]{"},
    stringstyle=\color{green!50!black},
    morestring=[s][\color{black}]{\ \ "}{":},
    keywordstyle=\color{blue},
    keywords={true,false,null},
    literate=
     *{0}{{{\color{red}0}}}{1}
      {1}{{{\color{red}1}}}{1}
      {2}{{{\color{red}2}}}{1}
      {3}{{{\color{red}3}}}{1}
      {4}{{{\color{red}4}}}{1}
      {5}{{{\color{red}5}}}{1}
      {6}{{{\color{red}6}}}{1}
      {7}{{{\color{red}7}}}{1}
      {8}{{{\color{red}8}}}{1}
      {9}{{{\color{red}9}}}{1}
      {.}{{{\color{red}.}}}{1}
      {:}{{{\color{gray}{:}}}}{1}
      {,}{{{\color{gray}{,}}}}{1}
      {\{}{{{\color{gray}{\{}}}}{1}
      {\}}{{{\color{gray}{\}}}}}{1}
      {[}{{{\color{gray}{[}}}}{1}
      {]}{{{\color{gray}{]}}}}{1},
}

\title{DACTYL: Diverse Adversarial Corpus of Texts Yielded from Large language models}

\author{ Shantanu Thorat\thanks{Work completed as master's thesis at University of Cambridge.} \\
	Department of Computer Science \& Technology\\
	University of Cambridge\\
	\And
	Andrew Caines \\
	Department of Computer Science \& Technology\\
	University of Cambridge\\}

\date{}

\renewcommand{\shorttitle}{DACTYL}

\hypersetup{
pdftitle={A template for the arxiv style},
pdfsubject={q-bio.NC, q-bio.QM},
pdfauthor={David S.~Hippocampus, Elias D.~Striatum},
pdfkeywords={First keyword, Second keyword, More},
}

\begin{document}
\maketitle

\begin{abstract}
AI-generated (AIG) texts are diffusing rapidly into our daily lives, blurring the lines between human and AI. Some use cases are malicious, such as generating misinformation on social media, producing fake reviews to mislead consumers, or writing false news articles. Thus, there emerges a need to build an AIG text detector. However, existing AIG text detectors struggle in real-world settings despite succeeding in internal testing, suggesting that they may not be robust enough. We rigorously examine the machine-learning procedure to build these detectors to address this. Most current AIG text detection datasets focus on zero-shot generations, but little work has been done on few-shot or one-shot generations, where LLMs are given human texts as an example. In response, we introduce the Diverse Adversarial Corpus of Texts Yielded from Language models (DACTYL), a challenging AIG text detection dataset focusing on one-shot/few-shot generations. We also include texts from domain-specific continued-pre-trained (CPT) language models, where we fully train all parameters using a memory-efficient optimization approach. Many existing AIG text detectors struggle significantly on our dataset, indicating a potential vulnerability to one-shot/few-shot and CPT-generated texts. We also train our own classifiers using two approaches: standard binary cross-entropy (BCE) optimization and a more recent approach, deep X-risk optimization (DXO). While BCE-trained classifiers marginally outperform DXO classifiers on the DACTYL test set, the latter excels on out-of-distribution (OOD) texts. In our mock deployment scenario in student essay detection with an OOD student essay dataset, the best DXO classifier outscored the best BCE-trained classifier by 50.56 macro-F1 score points at the lowest false positive rates for both. Our results indicate that DXO classifiers generalize better without overfitting to the test set. Our experiments highlight several areas of improvement for AIG text detectors, which can aid in the ongoing race between detectors and adversaries. 
\end{abstract}


\section{Introduction}
\label{firstcontentpage} 

\subsection{Overview}

AI-generated (AIG) text detection is the latest text classification task, a response to the flood of AIG texts in various domains, such as social media posts, reviews, and student essays. AIG text detection is a variation of authorship attribution --- a classifier determines if an LLM or a human wrote a text. We highlight several use cases of such a classifier:

\begin{itemize}
    \item Student essays --- Leading plagiarism detector Turnitin quickly deployed an AIG text detector to determine if students were passing off LLM-generated assignments as their own \citep{turnitinUnderstandingFalse}. 
    \item Fake review detection --- Roughly one in five Google Reviews may be AIG \citep{originalityFrom2019}.
    \item Pre-screening LLM training corpora --- \cite{shumailov2024ai} demonstrated that Meta's OPT-125M language model's text generation degraded after training on its own texts --- a phenomenon known as model collapse. The early generations of LLMs (pre-2022) did not have to worry about this issue, as individuals rarely employed LLMs. However, current LLM developers may have to start filtering out AIG texts in newer corpora, as LLM popularity explodes \citep{openai400million}.  
\end{itemize}

Unfortunately, AIG text detection has many challenges --- primarily that classifiers deployed in real-world settings struggle. For example, several universities have opted \textit{not} to use Turnitin's AI text detection tool due to many false positives \citep{vanderbiltGuidanceDetection}. Yet, classifiers are capable of achieving high performance on various AIG text datasets: two classifiers recently achieved at least a 99\% true positive rate at just a 5\% false positive rate on the Robust AI Detection (RAID) dataset \citep{dugan_raid_2024}. Thus, it appears that classifiers have the \textit{potential} to differentiate between human and AIG texts in controlled test settings, but it does not help them when facing AIG or human texts ``in the wild.''

This deployment gap inspires us to rigorously inspect traditional AIG text detector training. AIG text detection follows the conventional machine-learning workflow: construct a dataset, train a classifier, and evaluate on a test set (not used in training). Existing AIG datasets tend to include zero-shot (no example) LLM-generated text. We suspect that giving an LLM a human example would create a more human text, as LLMs are few-shot learners \citep{brown2020language}. Providing an LLM with a human example is also a reasonable technique an adversary might use to evade detection. Classifiers often train with binary cross-entropy to minimize the difference between the classifier's predicted scores (typically probability scores from 0 to 1) and the target label distribution \citep{pytorchBCELossx2014}. Evaluation is done with conventional binary classification metrics, which we explain later in Chapter 2. Given the strong performance on test sets but not in real-world deployment, we suspect that classifiers may overfit to the test set (which follows the training distribution more closely than thought).  

Our main contributions to the AIG text detection workflow are as follows. 
\begin{itemize}
    \item We introduce the Diverse Adversarial Corpus of Texts Yielded from Large language models (DACTYL) dataset, an AIG text detection dataset covering six domains and including one-shot/few-shot generated texts from 11 LLMs. Additionally, we provide an adversarial test set consisting of texts generated from 18 small language models (SLMs) which we have continued to pre-train (earlier works refer to this as fine-tuning \citep{salminen_creating_2022}).
    \item We train our classifiers on the DACTYL dataset using deep X-risk optimization. X-risks refer to a group of evaluation metrics that compare how well a classifier can contrast a positive sample (AIG text) with a negative sample (human text) \citep{yuan2023libauc}. Area Under the Curve (AUC) is a standard X-risk metric. 
    \item We evaluate our DACTYL-trained classifiers alongside existing AIG text detectors using the two-way partial AUC (Area Under the Curve) metric. We select this metric and three other X-risk metrics, as it is far less optimistic about classifier performance and highlights differences between classifiers better.  
    \item Finally, to ensure that our classifiers are generalizing, we evaluate them on an additional test set containing AIG texts from existing datasets to simulate an out-of-distribution (OOD) scenario to reflect real-world conditions better. 
\end{itemize}

We observe that pre-trained detectors struggle significantly on the DACTYL dataset, demonstrating a potential weakness in detection. Our continued pre-trained SLMs also degrade performance (compared to vanilla or ``out-of-the-box'' SLMs) --- even for DACTYL-trained classifiers. While deep X-risk optimized (DXO) classifiers do not perform as well as the binary cross-entropy (BCE) trained classifiers on the DACTYL test set, the DXO classifiers outperform the BCE classifiers on our OOD dataset. In our simulated deployment scenario for student essay detection, the BCE classifier's best false positive rate was 91.41\%  (85.72\% true positive rate), while the DXO classifier's lowest was just 0.66\% (84.16\% true positive rate). This performance gap starkly contrasts with the DACTYL student essay subset performance, where the BCE classifier marginally outperformed the DXO classifier. This example emphasizes the limitation of using traditional binary cross-entropy loss.

\subsection{Viability of AIG Text Detection}
We address a common claim about AIG text detectors in this section. AIG text detectors aren't perfect, and many researchers have strongly cautioned against using AIG text detectors, even suggesting the problem is impossible \citep{nicks_language_2024}. \cite{nicks_language_2024} argue that fine-tuning a language model to generate texts to bypass a detector is trivial. When a new detector is released, adversaries can reapply the same fine-tuning technique, becoming an endless cycle between adversaries and detectors.

While this is a legitimate concern, other fields have witnessed this ``arms race'', such as cybersecurity. Social media bot detection is an active race (known as bot evolution) dating back to 2010 \citep{cresci2020decade}. Yet, this has not dissuaded researchers in the field. Like defending against malicious bots, building an AIG text detector requires re-framing the solution --- the goal is not to create a \textit{one-time} detector but rather to train a detector robust enough for a (hopefully) considerable amount of time. Once an organization releases a classifier, it should continuously monitor it and attempt to identify its weaknesses before attackers exploit it. One AIG text detector, Pangram Labs, uses a similar approach: they identify difficult texts that confuse their detector and retrain on those \citep{emi-etal-2025-pangram}. Pangram Labs' continuous improvement technique boosts generalization: their detector outperforms all other pre-trained detectors by a considerable margin on the DACTYL test set. AIG text detection is an evolving problem, but the problem is not unsolvable. Instead, one should not expect a static solution to perform well over time.

\section{Background \& Related Work}

\subsection{Existing Work}
While the emergence of ChatGPT in late 2022 led to a surge in research in AI-generated (AIG) text detection, researchers have been analyzing this problem before its release. Early works focused on fake review or tweets detection using models such as GPT-2 \citep{adelani_generating_2020}  \citep{salminen_creating_2022} \citep{fagni_tweepfake_2021}. Earlier existing detection methods used classical machine learning models to distinguish between AIG and human-written texts. However, advancements in large language models (LLMs) and the transformer architecture have significantly improved text generation quality. Fortunately, more recent work has demonstrated the power of the transformer architecture for text classification --- including strong performance in AIG text detection.  

\subsection{Overview of Existing AIG Text Detection Datasets}

\subsubsection{Pre-ChatGPT Era: Fake News, Fake Reviews, and TweepFake}
\label{sec:prechatgptera}
The 2016 US election highlighted concerns regarding misinformation via fake news, prompting research into detecting fake human-written news. OpenAI's release of GPT-2 in 2019 marked a shift in research focus to investigating the malicious use of text generation models. Zellers et al. developed GROVER, a transformer-based model that can create and detect AIG news articles \citep{zellers_defending_2019}. The authors trained different-sized GROVER models on their news corpus (RealNews corpus). GROVER's smallest model (124M parameters) achieved a smaller perplexity (a measure of how surprising or unpredictable a text is to a model) in news generation than the largest GPT-2 model used (355M parameters): 15.2 vs 17.4. Regarding detection, fine-tuning GROVER for text classification generally outperformed BERT-Large and BERT-base classifiers \citep{zellers_defending_2019}. 

In 2020, \cite{adelani_generating_2020} demonstrated how to weaponize GPT-2 to mass produce fake Amazon and Yelp reviews effectively. The authors used a one-shot prompting technique that gave GPT-2 a ``seed review'' to generate reviews. To maintain a similar sentiment, they discarded generated reviews that didn't match the seed review's sentiment \citep{adelani_generating_2020}. The authors also fine-tuned a 117M GPT-2 model to improve generation quality as they noted that the base GPT-2 model's texts didn't resemble reviews. Detection efforts using GROVER, GLTR, and OpenAI's initial GPT-2 detector showcased that these classifiers outperformed humans in terms of equal error rate (EER --- i.e., the point where the false positive rate is equal to the false negative rate --- lower EERs are better). ``Fusing'' the detectors' scores together in an ensemble approach resulted in an EER of 22.5\%. 

\cite{salminen_creating_2022} later fine-tuned a 1.5B parameter model of GPT-2 on Amazon reviews. They trained their own RoBERTa classifier, fakeRoBERTa, which outperformed a classical machine learning approach, naive Bayes SVM by 2 macro-F1 score points (97 vs 95) \citep{salminen_creating_2022}. Their results also confirmed that humans are poor judges of distinguishing between human and AIG reviews (55\% accuracy vs. fakeRoBERTa's 96\%).  

In contrast to the datasets mentioned above, TweepFake is comprised mainly of AIG tweets posted by \textit{actual} Twitter bots from various models (such as GPT-2, RNNs, etc.). Some fine-tuned models generated the tweets \citep{fagni_tweepfake_2021}. The authors used 12,786 AIG tweets in their dataset. Fine-tuned transformer models such as BERT and RoBERTa dominated, but the authors highlighted that a GRU (gated recurrent unit) model trained on character embeddings marginally outperformed transformer models on detecting GPT-2-generated texts. 

\subsubsection{Large Language Model Era: MAGE and RAID}

Nearly all the datasets mentioned in section \ref{sec:prechatgptera} limited themselves to GPT-2 or one generative model family and exclusively focused on one domain. However, the two leading AIG text datasets, MAGE and RAID, include more variety in their texts --- in terms of domains and models. 

The MAGE (Machine-Generated) dataset, released in 2023 (previously called the DeepFake Text Detect dataset), spans seven domains and 27 different LLMs. MAGE evaluates the generalizability of classifiers: how well do classifiers perform on unseen models or domains? Once again, a fine-tuned transformer model, Longformer, outperformed various other classifiers. For example, the Longformer classifier achieved an AUC (Area Under the (Receiver Operating Characteristic) Curve, which equals the probability that a classifier assigns a higher score to the positive sample than the negative sample) score of 95 on the unseen models testbed. In contrast, the next best classifier (FastText) only had 74. However, Longformer struggled with paraphrased texts, with just an AUC score of 75. The main reason for the performance drop was the perplexity distribution shift in AIG texts after paraphrasing: AIG texts had increased perplexity, making them more similar to human texts. We note that \cite{li_mage_2024} did not include paraphrased texts in the original training set. 

Paraphrasing is just one common attack to bypass current detectors. The Robust AI Detection (RAID) dataset includes multiple domains like MAGE but also investigates performance on 11 adversarial attacks \citep{dugan_raid_2024}. RAID is one of the most exhaustive datasets with 6.2 million texts across the train and test splits, with an additional 2.3 million texts in an out-of-distribution testbed. The authors of RAID did not develop their classifiers but evaluated other classifiers on their dataset. At a COLING 2025 competition, Cross-Domain Machine-Generated Text Detection Challenge, which used RAID, the highest performing team used the distilRoBERTa-base model, an 82M parameter size model \citep{dugan_genai_2025}. distilRoBERTa-base's small size suggests that even small classifiers have the \textit{potential} to be robust to adversarial attacks (97.7\% TPR at 5\% FPR).

\subsection{Limitations of Existing Benchmarks}

Despite the exhaustive coverage of domains and adversarial attacks by MAGE and RAID, we highlight three significant limitations.

\textbf{Lack of fine-tuning/continued pre-training.} Nearly all pre-ChatGPT research involved detecting texts from fine-tuned models specific to the target domains (e.g., tweets or reviews). However, both MAGE and RAID only focus on pre-trained models. Existing work demonstrates that fine-tuned LLMs can evade existing detectors relatively easily \citep{nicks_language_2024}. Fine-tuned LLMs generally show improvements over the base models, making them ideal as an adversarial attack to bypass detection. 

We note that some papers (such as \cite{salminen_creating_2022}) refer to their domain-specific models as ``fine-tuned.'' However, they ``fine-tune'' on a raw corpus, which is technically considered continued pre-training (CPT) \citep{databricksCharacterizingDatasets}. CPT involves further training a pre-trained LLM on a domain-specific corpus with no additional tasks --- just next token prediction. Fine-tuning typically involves tasks that require an input to map to an output, such as question answering. 

\textbf{One-shot/Few-shot prompting exploration.} MAGE explores three categories of prompts: continuation, topical (respond to the prompt), and specified (respond to the prompt given additional context). RAID exclusively focuses on zero-shot prompting, arguing that providing examples to LLMs does not align with real-world situations \citep{dugan_raid_2024}. However, detectors may struggle to identify such one-shot generations, leading to possible exploitation. 

\textbf{Static benchmarks.} Static datasets are relatively common in academic settings; however, real-world distribution shifts are standard and can limit the effectiveness of future deployments trained on a single dataset. These shifts in AIG text detection can include new attacks or even new LLM introductions. Rather than regenerating multiple datasets from scratch, it is more convenient to keep updating a single dataset --- similar to version control software. This technique allows model developers to see how AIG-text detection (and evasion) evolves. 
\subsection{Detecting AIG Texts}

Constructing a dataset is the first part of developing an AIG text detector. However, training the classifier is another challenge. We highlight two types of classifiers: transformer-based and the less common but notable zero-shot classifiers. 
\subsubsection{Transformer-Based}
Existing works have demonstrated the superiority of transformer-based models over traditional methods. These are often pre-trained language models such as BERT (Bidirectional Encoder Representations with Transformers) with a classification head (layer) on top \citep{devlin-etal-2019-bert}. Unlike the GPT family, BERT can exploit contexts from both left-hand and right-hand sides, enabling powerful performance on tasks such as text classification. GPT can only take in context from the left-hand side (i.e., when generating the next token, GPT will only have access to all tokens that came before it, or the left-hand side tokens). BERT is often pre-trained on massive text corpora using masked language modeling (MLM): BERT has to ``fill in the blank'' for missing words. Additionally, it uses next-sentence prediction in training. After pre-training, BERT and its variants can be fine-tuned for AIG text classification. We list some of the popular variants. 

\textbf{BERT-tiny.} The smallest within the original family, BERT-tiny has only 4.4 million parameters with two layers and a hidden size of 128 \citep{bhargava2021generalization} \citep{DBLP:journals/corr/abs-1908-08962}.

\textbf{TinyBERT.} Not to be confused with the BERT-tiny, TinyBERT was developed by researchers at Huawei Noah's Ark Lab. This model has around 14.5 million parameters \citep{jiao2019tinybert}. Jiao et al.\ used knowledge distillation (KD) to train TinyBERT using a larger BERT model. KD involves training the student model (TinyBERT) to mimic the teacher model's (BERT) behavior.

\textbf{distilRoBERTa-base.} Using knowledge distillation similar to TinyBERT, distilRoBERTa-base is the distilled version of RoBERTa-base. RoBERTa (Robustly Optimized BERT pre-training Approach) is a BERT model with a modified pre-training approach. Liu et al. pre-trained BERT with more data over longer sequences and updates (i.e., more training time) \citep{liu2019roberta}. The distilled version has around 82.8 million parameters. 

\textbf{ModernBERT-base.} ModernBERT-base involves significant architectural changes. Arguably, its distinguishing trait is that it uses RoPE (rotary positional embeddings), allowing it to scale to longer texts \citep{modernbert}. ModernBERT-base is also optimized for memory and speed, making it ideal for larger datasets. ModernBERT-base's performance is competitive with much larger models, such as DeBERTaV3-large (149 million parameters vs 434 million parameters) \citep{modernbert}. 

\textbf{DeBERTaV3-large.} Boasting 434 million parameters, DeBERTaV3-large is one of the largest variations of BERT. DeBERTaV3 does not use MLM during its pre-training stage; rather, it uses RTD (replaced token detection), \ where DeBERTaV3 has to predict if a given token was replaced (corrupted) \citep{he2021debertav3}. 

\subsubsection{Zero-shot Classifier}
Zero-shot classifiers benefit from not needing training data. In contrast with transformer-based approaches, zero-shot detection has not been extensively explored, but we investigate one zero-shot classifier with promising results: the Binoculars approach \citep{hans_spotting_2024}. 

Binoculars exploits the perplexity gap between AIG texts and human-written texts to distinguish between the two --- AIG texts have lower perplexity scores and are more predictable. However, perplexity alone is not a reliable signal; hand-crafted prompts can arbitrarily increase perplexity by stringing words that rarely appear together. For example, Hans et al. use a GPT-4 generated response with the prompt ``Can you write a few sentences about a capybara that is an astrophysicist?'' \citep{hans_spotting_2024}. The authors reported that AIG text detectors GPTZero and DetectGPT failed to identify the response as AIG. The words ``capybara'' and ``astrophysicist'' rarely appear together, which leads to a higher perplexity and can cause misclassification of AIG texts. 

To address the ``capybara'' problem, the authors propose a new metric called cross-perplexity. First, they compute the log perplexity for a given string $s$, LLM $M$, and $M$'s tokenizer $T$. $T$ can tokenize $s$ into a list of tokens denoted by integer values $x$ and $x_i$ is the $i$-th token in $x$.

\begin{equation}
   M(x) = Y
\end{equation}
$Y$ represents the probability distribution over the vocabulary $V$ such that for all tokens up to token $i-1$, the probability of token $j$ is equal to \citep{hans_spotting_2024}:
\begin{equation}
     Y_{ij} = P(v_j|x_{1:i-1})
\end{equation}
They define log-perplexity as: 
\begin{equation}
    \log{\text{PPL}_{M}(s)} = -\frac{1}{L}\sum_{i=1}^{L}\log{(Y_{ix_i})}
\end{equation}
Cross-perplexity involves computing perplexity across two LLMs to determine if one LLM's output is ``surprising'' to another's. They compute it for two models $M_1$ and $M_2$, given string $s$, as:
\begin{equation}
     \log{\text{X-PPL}}_{M_1,M_2}(s) = - \frac{1}{L} \sum_{i = 1}^L M_1(s)_i \cdot \log\left(M_2(s)_i\right).
\end{equation}
The cross-perplexity is the mean cross-entropy per token between $M_1$ and $M_2$'s outputs. 
Finally, the Binoculars score is defined as \citep{hans_spotting_2024}:
\begin{equation}
    \frac{\log{\text{PPL}_{M_1}(s)}}{\log{\text{X-PPL}_{M_1, M_2}(s)}}
\end{equation}
 The authors divide by cross-perplexity to normalize log-perplexity, which provides a metric that is ``invariant to the prompt'' \citep{hans_spotting_2024}. We expect a human-written text to be surprising to $M_1$ (numerator), but $M_2$'s outputs will not be as surprising to $M_1$'s (denominator). Because the Binoculars method scores human texts above AIG texts, the Binoculars scores must be multiplied by $-1$ to flip the ordering (that way, AIG texts get higher scores) \citep{hans_spotting_2024}.
 
\subsection{Evaluating Classifier Performance}
\label{sec:eval-classifier-performance}
\begin{figure}[h!]
    \centering
    \includegraphics[width=0.9\linewidth]{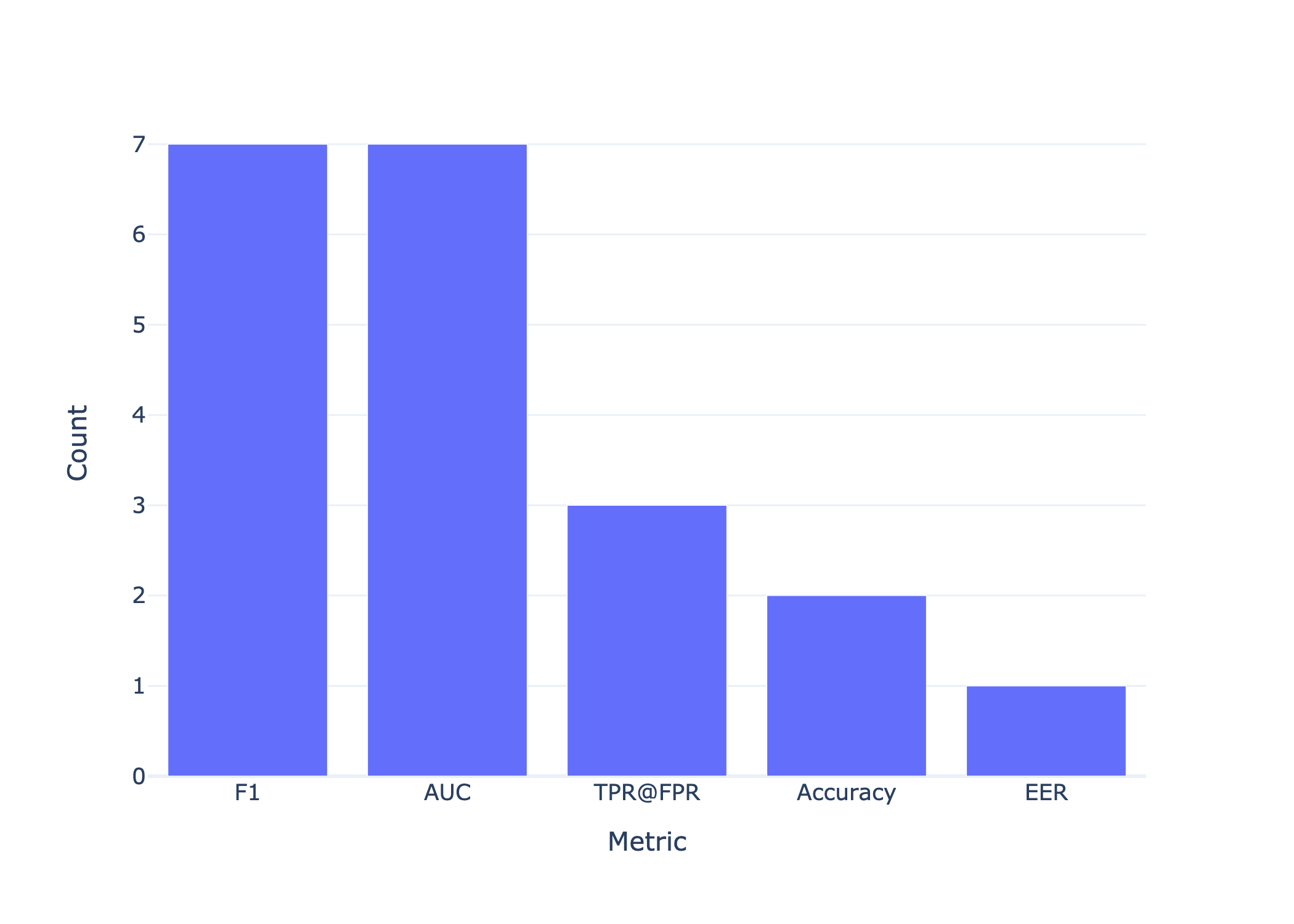}
    \caption{Primary metric reported across 20 AIG text detection papers \citep{cornelius-etal-2024-bust} \citep{bitgritGeneratedText} \citep{webisCLEF2025} \citep{guo-etal-2023-hc3} \citep{verma2023ghostbuster} \citep{macko2024multisocial}. }
    \label{fig:metric-dist}
\end{figure}

Figure \ref{fig:metric-dist} demonstrates the wide variety of metrics used to evaluate AIG text detection performance. However, there has not been much consistency in evaluation, making it challenging to compare classifiers and models over time. Macro-F1 and AUC scores are popular, but they have several disadvantages. 

Macro-F1 score, the harmonic mean between precision and recall, takes in binary predictions for data points and the true labels (e.g., no score for those labels). Because of this definition, threshold calibration must be done to get actual predictions. Deciding a threshold for AIG text detection is highly domain-dependent: LLM developers filtering out AIG texts in their corpora may tolerate lower thresholds at the expense of increased false positive rates (more human texts flagged as AIG). In contrast, plagiarism detectors such as Turnitin may prefer a higher threshold to minimize false positives. We also lose information about the scores themselves. 

While AUC does consider classifier performance across various thresholds, unlike the F1-score, it ignores precision --- a metric that assesses a classifier's correctness when it flags a text as AIG. The precision-recall curve (PRC) visualizes the trade-off between precision and recall (i.e., the TPR). We can approximate the area under PRC (AUPRC) via the Average Precision (AP) score. The AP score's distinguishing factor is penalizing high-scoring false positives (in our case, human texts with an extremely high AIG text score) \citep{mcdermott2024closer}. This behavior is beneficial in student essay detection, as educators can easily lose trust in a classifier with over-confident predictions. 

Another drawback of AUC is the presence of high FPR values along the ROC curve, which inflates the score. Typically, thresholds with high FPRs and low TPRs aren't helpful (e.g., the lower right quadrant of the ROC curve). Thus, it is intuitive to set a limit on the FPR and focus on the area under the curve until that limit. This metric is one-way partial AUC or pAUC. We can restrict partial AUC even further: two-way partial AUC (tpAUC) focuses on estimating AUC by giving a lower bound for TPR and an upper bound for FPR, denoted by $\alpha$ and $\beta$, respectively. 

When evaluating classifiers on our dataset, we report the following metrics: tpAUC($\alpha$,$\beta$), pAUC($\beta$), AP, and AUC scores. Together, these metrics can highlight weaknesses in classifier performance while considering the classifier's scores. We provide a ranking criterion across these measures used to compare classifier performance (either between classifiers or evaluating an individual classifier within a dataset on specific factors):

\begin{enumerate}
    \item tpAUC --- The most restrictive metric out of the four that prioritizes performance such that $\alpha \leq$ TPR and $\beta \geq$ FPR. A classifier can achieve an AUC $\geq$ 90 but still have a significantly smaller tpAUC.
    
    \item pAUC --- The second most restrictive metric that prioritizes performance such that FPR $\leq \beta$.

    \item AUC
    \item AP 
    
\end{enumerate}

We use AUC as the third tiebreaker rather than the AP score since the AUC values improvements across all of a classifier's predictions rather than focusing on high-scoring mistakes. Also, for AIG detection, there may be some cases where false negatives (AIG texts classified as human) should be minimized; these mistakes are often low-scoring, which the AP score doesn't prioritize as heavily \citep{mcdermott2024closer}. However, it is rare in practice to reach the second or third tiebreaker unless $\alpha$ and $\beta$ are too strict. 

\begin{figure}[h!]
    \centering
    \includegraphics[width=0.6\linewidth]{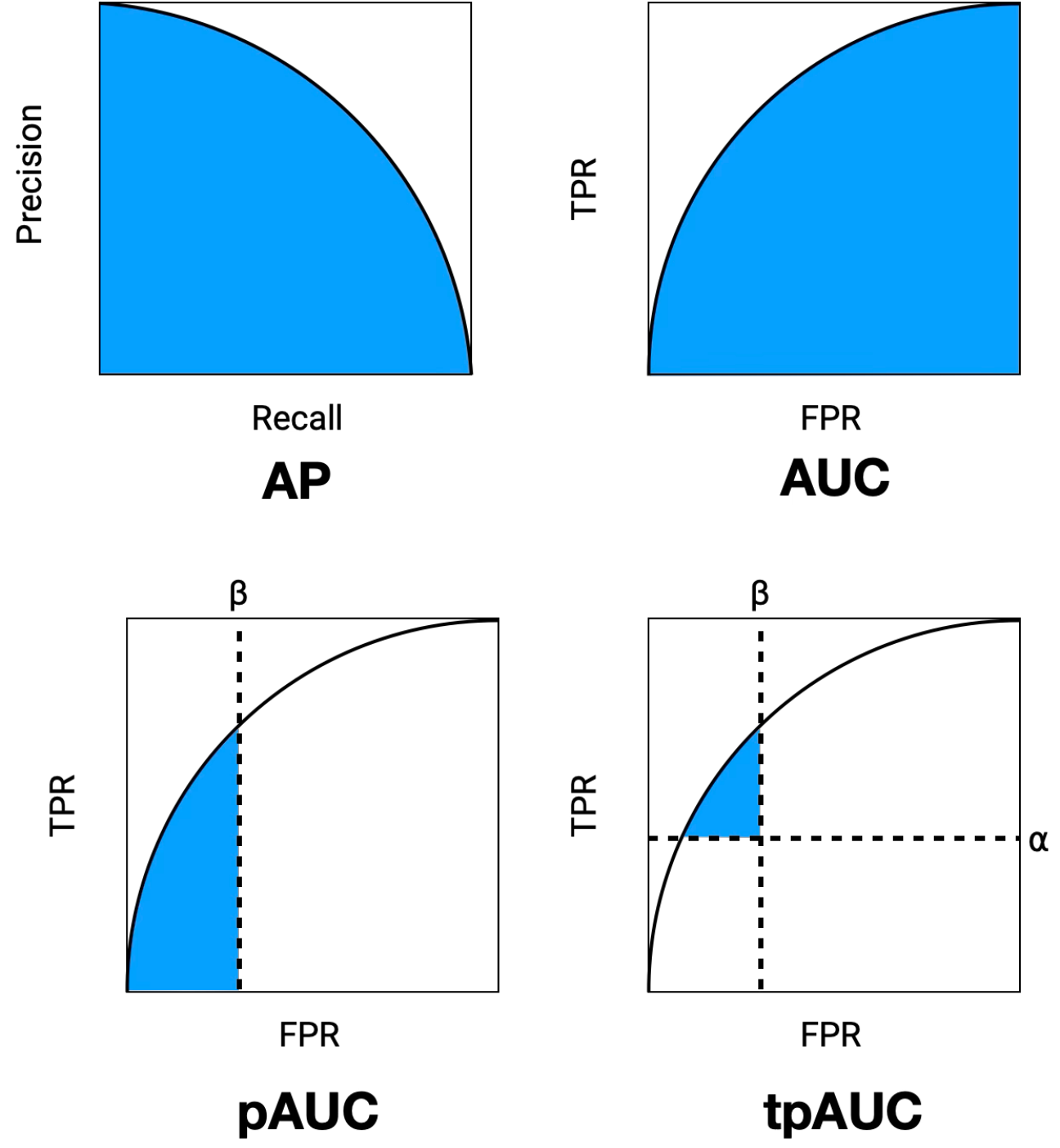}
    \caption{Comparison of the four X-risks used to evaluate classifier performance.}
    \label{fig:x-risks-used}
\end{figure}

These four metrics are part of a family of measures called X-risks, which involve contrasting data points. We calculate tpAUC using the LibAUC library, and scikit-learn for the remaining three \citep{yuan2023libauc} \citep{pedregosa2011scikit}. X-risk measures can be optimized explicitly, as we see in the following section.   

\subsection{Deep X-risk Optimization}
Regarding classifier performance, most text classifiers use cross-entropy loss or its variants during training; HuggingFace's \verb|transformers| library's default loss function for classification is the \verb|CrossEntropyLoss| function. However, vanilla cross-entropy loss is not robust to class distribution differences. Deep X-risk optimization (DXO) has demonstrated improvements in imbalanced classification across multiple domains, such as image and graph classification. For example, the top-scoring teams at MIT's AICures challenge utilized AUC and AP maximization to achieve first place \citep{wang_advanced_2022}. DXO algorithms also have several ideal properties, such as guaranteeing theoretical convergence no matter the mini-batch size and scaling well for large datasets \citep{yuan2023libauc}. Empirically, these methods perform well on domains that feature out-of-distribution samples (fourth place on the OGBG-MOLHIV benchmark). These properties are crucial for AIG detectors: AIG text datasets are relatively large, but compute resources may be limited (hence, smaller batch sizes may be preferred). Robust out-of-distribution performance is a necessary property for text classifiers: the number of combinations of strings is too large to obtain a \textit{complete} representative training set --- in AIG text classification, this includes unseen models and unseen attacks.  

The LibAUC library provides DXO loss functions and optimizers for AUC, AP, and pAUC. Existing work has already explored AUC maximization in this field \citep{thorat_which_2024}. However, AUC maximization does not guarantee pAUC maximization (by extension, tpAUC) \citep{yuan2023libauc}. Thus, since our primary evaluation metric is tpAUC, we focus on the loss function for tpAUC. 

\subsection{pAUC Optimization}

Partial AUC optimization relies on distributional robust optimization (DRO). Given a set of loss functions ($\ell_{1}, \ell_{2},...\ell_{n}$) We can write a DRO loss as \citep{zhu2022auc}: 
\begin{equation}
    \hat{L}_{\phi}(\cdot) = \max_{\mathbf{p} \in \Delta}{\sum_{j}{p_j\ell_j(\cdot)} - \lambda D_{\phi}(\mathbf{p} , 1/n)}
\end{equation}
where 
\begin{equation}
\label{eqn:divergence-formula}
    D_\phi(\boldsymbol{p}, 1/n) = \frac{1}{n}\sum_{i}{\phi(np_i)}.
\end{equation}

$\lambda$ is a positive parameter, $D_{\phi}$ is a divergence measure, $p_j$ is a weight parameter, and $\phi(\cdot)$ can either be the Kullback–Leibler (KL) divergence term or the conditional-value-at-risk (CVaR) term. We focus on the KL term as the tpAUC uses KL.  

\begin{equation}
    \phi_{kl}(t) = t\log{t} - t + 1
\end{equation}
Thus, equation \ref{eqn:divergence-formula} can be rewritten as:
\begin{equation}
      D_\phi(\mathbf{p}, 1/n) = \sum_{i}{p_i\log(np_i)}.
\end{equation}
tpAUC is an extension of pAUC. pAUC defines a loss function for each positive sample ($x_i$) as
\begin{equation}
    \hat{L}_{\phi}(\mathrm{w}; x_i) = \max_{\mathbf{p} \in \Delta}{\sum_{j}{p_jL(\mathrm{w}; x_i)} - \lambda D_{\phi}(\mathbf{p}, n)},
\end{equation}
where $L$ is a surrogate loss function (e.g., squared-hinge loss):
\begin{equation}
    L(\mathbf{w}; x_i, x_j) = \ell(h_{\mathbf{w}}(x_i) -h_{\mathbf{w}}(x_j)).
\end{equation}
For pAUC, this means the objective function is \citep{zhu2022auc}:
\begin{equation}
    \min_{\mathbf{w}}\frac{1}{n_+}\sum_{\mathbf{x}_i \in \mathbf{S}_+}\lambda \log\frac{1}{n_-}\sum_{\mathbf{x}_j \in \mathbf{S}_-}\exp\left(\frac{L(\mathbf{w}; \mathbf{x}_i, \mathbf{x}_j)}{\lambda}\right)
\end{equation}
where $n_+$ and $n_-$ are the sample sizes of the positive and negative samples (sets denoted as $
\mathbf{S}$), respectively. 

We can extend this to tpAUC as

\begin{equation}
    \min_{\mathbf{w}}\lambda'\log \frac{1}{n_+}\sum_{x_i \in \mathbf{S_+}}\left(\frac{1}{n_-} \sum_{x_j \in \mathbf{S_-}}\exp(\frac{L(\mathbf w; \mathbf x_i,\mathbf x_j)}{\lambda})\right)^{\frac{\lambda}{\lambda'}}
\end{equation}

where $\lambda$ and $\lambda'$ are parameters that can be tuned to set the range of TPR and FPR, respectively. 

\subsection{APOLLO Optimizer}
The APOLLO (AProximated gradient scaling for memOry efficient LLm Optimization) optimizer offers ``SGD-like memory'' with ``AdamW-level performance'' \citep{zhu2024apollo}. The APOLLO optimizer is designed explicitly for \textit{full} fine-tuning or pre-training. AdamW has become the de facto standard for fine-tuning, but its memory usage is high, even for smaller-scale models, with a 7B model requiring 58 GB of GPU memory. AdamW can be optimized further by restructuring its learning rate update rule such that each channel or tensor shares the exact gradient scaling factor \citep{zhu2024apollo}. To achieve this scaling, Zhu et al.\ used an auxiliary optimizer state to compress gradient information (i.e., exploiting low-rank properties). The APOLLO optimizer goes further in memory savings by substituting the Singular Value Decomposition (SVD)-based (low-rank) projections with random projections. A low rank of 1 is enough for the APOLLO optimizer to work with --- this extreme version is called APOLLO-Mini. For reference, the APOLLO-Mini optimizer only needs under 20 GB of GPU for a 7B model. 

The APOLLO optimizer's efficiency makes it an ideal tool for adversaries limited by computing resources. Combined with the many publicly available 1 to 7 billion parameter language models on HuggingFace, we can expect more individuals to fine-tune these models for nefarious purposes.

\section{Dataset Design}
\subsection{Overview}
We construct our dataset using eleven LLMs across six vulnerable domains. We source human texts from these domains using publicly available datasets. 

\subsubsection{Selected Domains}

We focus on these domains given two criteria: (1) their \textit{potential} for AIG text abuse and (2) existing evidence of abuse by LLMs.  

\subsubsubsection{Tweets}
X, formerly known as Twitter, is a microblogging platform that enables users to publish very short (less than 300 characters) posts or tweets. Former US president Barack Obama used the platform extensively during his 2008 and 2012 campaigns to his success. However, the 2016 US presidential election witnessed a surge in voter manipulation ranging from misinformation to trolls \citep{roeder_why_2018}. The potential impacts of generative AI on automating disinformation campaigns were of particular concern in the 2024 elections in the US, UK, and EU \citep{heikkila_ai-generated_2024}. Fortunately, early evidence suggests that AI disinformation does \textit{not} appear to sway public opinions, according to reports from the Alan Turing Institute \citep{stockwell_ai-enabled_nodate}. However, generative AI may no longer be used to persuade others but to cause confusion and chaos \citep{heikkila_ai-generated_2024}. Thus, detecting AIG texts in social media remains critical. Additionally, AIG text detectors seem to struggle with shorter texts; a popular AIG text detector, GPTZero, only checks texts with at least 250 characters, but tweets usually have substantially fewer characters.   

Given electoral concerns, we use FiveThirtyEight's 3 Million Russian Troll tweets dataset. University of Clemson researchers Linvill and Warren assembled this Twitter corpus, with tweets from 2012 to 2018 (a majority of the tweets were published between 2015 and 2018) \citep{roeder_why_2018}. The Internet Research Agency employed 400 employees in their troll factory (an organization dedicated to creating troll accounts and activity) targeting the 2016 US election. \cite{linvill2020troll} categorized the troll tweets into five major categories: 
\begin{itemize}
    \item News feeds: These tweets aggregate local news, often from legitimate sources \citep{roeder_why_2018}.
    \item Hashtag games: Tweets that participated in Twitter hashtag trends.
    \item Left-wing: These tweets attempted to divide the Democratic Party base and encourage low voter turnout.
    \item Right-wing: These tweets act like typical supporters of the Republican Party. 
    \item Fearmonger: These tweets spread fake news about a fictional crisis.
\end{itemize}
 Out of these five, we focused on left-wing, right-wing, and fearmonger tweets as these tweets are more likely to stoke division and chaos. Tweets sampled from this dataset are classified as human as the Internet Research Agency employed a substantial number of humans, and most tweets come before any widespread advancement of generative text technology. 
 
\subsubsubsection{Reviews}
Product reviews are another domain of particular interest. Fake reviews have been a problem long before the introduction of LLMs, but increased online shopping during the COVID-19 pandemic led to a surge in fake reviews \citep{mccluskey_inside_2022}. By July 2023, users reported seeing fake AIG reviews on platforms such as Amazon \citep{probert_online_2023}.  In late 2024, the US Federal Trade Commission (FTC) criminalized the purchasing and distribution of fake reviews (including those written by humans or LLMs) by businesses \citep{federal_trade_commission_federal_2024}. The danger of fake reviews is their influence on consumers: a consumer watchdog organization, the Transparency Company, estimates that these reviews influence around 300 billion USD in consumer spending in the US alone \citep{cavazos_high_2024}. In that report, they saw around 2.3 million AIG reviews using Pangram Lab's detector, out of 73 million reviews \citep{haduro_internet_2024} \citep{cavazos_high_2024}.

For our dataset, we focused on Amazon reviews, as existing datasets typically focused on Yelp reviews \citep{li_mage_2024} \citep{dugan_raid_2024}. We use Hou et al.'s Amazon 2023 review dataset \citep{hou_bridging_2024} as our ``human'' dataset. Although there is a chance of possible data contamination from AIG reviews, fake AIG reviews did not appear in large numbers until June 2023 \citep{cavazos_high_2024} \citep{probert_online_2023}. The latest Amazon review from our dataset has a timestamp of April 20th, 2023. 

\subsubsubsection{Abstracts}

Abstracts and academic papers are also prone to LLM abuse. Around two-thirds of sampled papers from Google Scholar had signs of AIG text in them (e.g., ``as of my last knowledge update''), according to a Harvard Kennedy School paper \citep{haider_gpt-fabricated_2024}. Around 57\% of those papers targeted fields that heavily influence public policy, such as health, environment, and computer science. The long-term ramifications of AIG publications in the wild include loss of public trust in academia and misleading policy-makers \citep{haider_gpt-fabricated_2024}.

To address this concern, we focus on abstracts, arguably one of the most essential parts of modern publications: many potential readers will read the abstract only to gain a first impression of the paper \citep{alspach2017writing}. In a similar pattern to tweets, abstracts are much shorter than their papers; thus, many detectors may struggle to detect AIG  abstracts. We use arXiv's abstract dataset as the source for human abstracts \citep{arxiv_org_submitters_2024}. We selected abstracts with papers that belong to a single category (i.e., subject); this allows us to examine difficulties in AIG text detection across categories without dealing with multi-category papers. Since arXiv explicitly allows for AIG content (as long as the authors declare it), we only sample abstracts that were updated on or before the release of ChatGPT (November 30th, 2022) \citep{noauthor_arxiv_policy_update_2023} \citep{acres_chatgpt_2023}. 

\subsubsubsection{News}
Similar to tweets, concerns about fake news also emerged during the 2016 US election. Generative AI is now automating the production of these false articles: NewsGuard reported that from May 2023 to December 2023, the number of sites hosting AIG news articles exploded from 49 to over 600 \citep{verma_rise_2023}. AIG news combined with other generative AI methods (e.g., deepfakes) can cause confusion among voters during elections. 

To address potential disinformation abuse, we use the Information Security and Object Technology (ISOT) research lab's fake news dataset \citep{ahmed_detection_2017} \citep{ahmed_detecting_2018}. \cite{ahmed_detecting_2018} constructed this dataset before 2018, and thus should not have AIG articles. The original purpose of this dataset was to train classifiers to distinguish between fake and real news articles across various topics. 

\subsubsubsection{Student Essays}
Academic cheating and plagiarism abuse have sparked interest in efforts to detect LLM usage in student essays. Existing detectors seem to struggle with this domain; in its beta release, Turnitin's AI text detector faced issues in real-world settings \citep{scarfe2024real}. For example, early versions of these detectors were biased against non-native English speakers \citep{liang2023gpt}. Several universities have ultimately opted out of using AI text detectors \citep{ghaffary_universities_2023}. However, relying on humans to discern between AI and authentic texts is arguably worse: in a live experiment at the University of Reading, 94\% of AIG submissions bypassed human detection \citep{scarfe2024real}. At the time of the study, summer of 2023, the university did not have an AI text detector, so markers were given vague signs to look out for in student essays. Exacerbating the issue, AI submissions generally scored higher than their human counterparts \citep{scarfe2024real}. 

In creating this domain's texts, we sampled human essays from two datasets: the Ivy Panda dataset and the English Language Learner Insight, Proficiency and Skills Evaluation (ELLIPSE) Corpus \citep{vechtomov_qwedsacfivypanda-essays_nodate} \citep{crossley_english_2023}. \cite{vechtomov_qwedsacfivypanda-essays_nodate} scraped essays from the Ivy Panda website containing numerous examples. The editorial team verifies essays to ensure they are human-written; however, they do not give any details about the verification process \citep{ivypandaFirstMoverAdvantages}. We selected ELLIPSE in response to concerns that AI text detectors are biased against non-native English speakers. The corpus contains student essays written for national and state-wide standardized exams in the US \citep{crossley_english_2023}.

\subsubsubsection{Creative Writing}

Generative AI has quietly impacted the creative writing field since 2022. In 2023, Amazon's Kindle Direct Publishing implemented a publishing limit of three books per day in response to a surge of AIG books. Additionally, they mandate authors to disclose AIG content in their books \citep{creamer_amazon_2023}. However, it is not clear how Amazon would enforce such a requirement. 

We source human texts and writing prompts from Reddit's WritingPrompts, an online forum for creative writing \citep{fan-etal-2018-hierarchical}. The authors constructed this dataset at some point in 2018. Thus, it is highly unlikely for AIG text contamination. 

\subsection{LLMs}
We generate our initial dataset from eleven LLMs (we specify parameter counts for models where that information is publicly available): 
\begin{itemize}
    \item OpenAI's GPT-4o-mini \citep{openaiGPT4oMini} and GPT-4o \citep{hurst2024gpt}
    \item Anthropic's Claude Haiku and Sonnet 3.5 \citep{Claude3S} \citep{anthropicIntroducingComputer}
    \item Mistral Small (24B)\citep{mistralMistralSmall} and Large 2 (123B) \citep{mistralLargeEnough}
    \item Google's Gemini 1.5 Flash and Pro \citep{team2024gemini}
    \item Meta's Llama 3.2 90B \citep{metaLlama32} and 3.3 70B \citep{metaLlama33}
    \item DeepSeek-V3 (671B) \citep{liu2024deepseek}
\end{itemize}

We included the OpenAI family due to its popularity --- the company reported it has around 400 million active users weekly in early 2025 \citep{openai400million}. We included the Anthropic models due to these models showing improvement over previous iterations in tasks such as instruction following (helpful for few-shot prompting) and creative writing \citep{AnthropicModelCA}. The Mistral models also demonstrate strong performance in instruction following, particularly the Large 2 model \citep{mistralLargeEnough}. The Gemini family shows improvements in instruction following compared to its previous versions, similar to the Anthropic family \citep{team2024gemini}. We selected Llama 3.2 90B due to its image and text support. While we do not utilize images as input, we are interested in seeing if this model architecture difference contributes to any detection difficulties. We included the 3.3 70B model as it ``provides enhanced performance relative to Llama 3.1 70B–and to Llama 3.2 90B when used for text-only applications'', according to Amazon Web Services \citep{amazonMetasLlama}. DeepSeek-V3 was not included in the original ten LLMs to construct the DACTYL dataset as it was unavailable until December 2024. However, given DeepSeek-V3's popularity, we included it as the eleventh model \citep{openai400million}. 

We accessed the OpenAI, Anthropic, Mistral, and Gemini models directly from the model developers' REST and Python APIs. We accessed Llama models through the AWS Bedrock API \citep{amazonAmazonBedrock}. We generated DeepSeek-V3 texts through the DeepInfra and Fireworks AI APIs \citep{deepinfraMachineLearning} \citep{fireworksFireworksDeveloper}.

We randomly sampled temperature and top-$p$ values from a uniform distribution $[0, 1]$. For Gemini and OpenAI models, we sampled temperature from a uniform distribution of $[0, 2]$, as suggested by the technical documentation \citep{openai_openai_2025} \citep{google_experiment_2025}. The temperature parameter dictates the randomness of the output; higher temperature values encourage randomness. Top-$p$, sometimes referred to as nucleus sampling, indicates to the LLM to only consider the top $p\%$ of candidate tokens when generating the next token. For example, a top-$p$ value of 0.2 means the LLM will only consider the top 20\% of tokens.

\subsection{Non-Adversarial Texts}

\subsubsection{Tweets}

To generate our AIG tweets, we used five-shot prompting: for a given prompt, we randomly sampled five tweets from one of the three troll categories to use as an example. For example, in code block \ref{listing:tweet-prompt}, we passed in a system prompt that gives background context about how the LLM should respond, instructing Gemini 1.5 Pro to use the user's inputted tweets (denoted by \verb|"role": "user"|) as an example before responding with its tweet (response in code block \ref{listing:tweet-response}).  We generated 500 training tweets, 200 validation tweets, and 200 testing tweets for each LLM and troll category. To avoid test leakage, we only sampled human tweets from the respective split to use as examples for prompting (i.e., only use human testing tweets as examples when prompting for AIG testing tweets). We report the distribution across human, AI, and troll categories in Table \ref{tab:tweet-dist}. 

\begin{table}[h!]
\centering
\caption{Distribution of human and AIG tweets.}
\begin{tabular}{@{}ccccccc@{}}
\toprule
\textbf{}               & \multicolumn{2}{c}{\textbf{Training}} & \multicolumn{2}{c}{\textbf{Validation}} & \multicolumn{2}{c}{\textbf{Testing}} \\ \midrule
\textbf{Troll Category} & \textbf{Human}      & \textbf{AI}     & \textbf{Human}       & \textbf{AI}      & \textbf{Human}     & \textbf{AI}     \\ \midrule
Left                    & 25465               & 5500            & 3172                 & 2200             & 3153               & 2200            \\
Right                   & 27186               & 5500            & 3388                 & 2200             & 3404               & 2200            \\
Fearmonger              & 4150                & 5500            & 506                  & 2200             & 523                & 2200            \\ \bottomrule
\end{tabular}
\label{tab:tweet-dist}
\end{table}

We originally sampled 86,000 human tweets across all splits (intended to be 68800/8600/8600 split). However, we observed that many LLM prompts we passed in got rejected --- the LLMs refused to generate a tweet provided in the examples due to their safety training. A substantial proportion of the tweets sampled during the five-shot prompting were toxic. To minimize this, we filtered out toxic tweets using the \verb|detoxify| library's toxic comment classifier (the ``unbiased'' model) \citep{Detoxify}. We removed any human tweets with a toxicity score greater than 0.1 (1 being highly toxic) from our initial sample of 86,000 tweets.

\begin{lstlisting}[float, language=json,caption=Example prompt used for Gemini-1.5-Pro (fearmonger troll). System prompts provide background context while user prompts are directly passed to the LLM.,label=listing:tweet-prompt]
[
  {
    "content": "You are a Twitter bot. You will be given example tweets, and then you have to generate new tweets in that same style. Generate only one tweet at a time.",
    "role": "system"
  },
  {
    "content": "Real Housewives of New York City star Kristen Taekman husband apologizes ...",
    "role": "user"
  },
  {
    "content": "wooow  that's not funny guys  #kochfarms https://t.co/zYfYxbsgKT",
    "role": "user"
  },
  {
    "content": "#DogThanking Where did you buy your #thanksgiving #turkey? https://t.co/OpnXHYSF3B",
    "role": "user"
  },
  {
    "content": "Investigation into 3 Washington firefighter deaths will follow more sensitive ...",
    "role": "user"
  },
  {
    "content": "'@NYCOER #KochFarms made a #turkey which makes people sick... Be careful my friends https://t.co/89MJynmpbx'",
    "role": "user"
  }
]
\end{lstlisting}

\begin{lstlisting}[float, language=json,caption=Response from code listing \ref{listing:tweet-prompt}, label=listing:tweet-response]
{
    "content": "Ewww #KochFarms #turkey is nasty!  Don't buy it!  #foodpoisoning"
}
\end{lstlisting}

\subsubsection{Reviews}
We used one-shot prompting for this domain (and the remaining domains): give the LLM a human review and prompt it to generate another review in that same style. We provide the LLM with the desired star rating $r$ and product information (product from category $p_c$) in the system prompt for further context. The example human review will have the same rating $r$ and be from the same category $p_c$. Each LLM generated 1000 training texts and 250 validation and testing texts. We report the AIG review distribution among product categories and star ratings in Table \ref{tab:review-dist}. We only generated reviews from 9 of the 34  categories with product information (metadata) stored in the condensed \verb|.parquet| format to save costs and time.

\begin{table}[h!]
\caption{DACTYL's AIG review distribution aggregated across all splits.}
\centering
\begin{tabular}{@{}cccccc@{}}
\toprule
                            & \multicolumn{5}{c}{Star Rating}                                               \\ \midrule
Product Category            & 1             & 2             & 3             & 4             & 5             \\ \midrule
All Beauty                  & 398           & 404           & 390           & 377           & 377           \\
Arts Crafts And Sewing      & 364           & 381           & 386           & 383           & 396           \\
Cell Phones And Accessories & 406           & 407           & 408           & 388           & 417           \\
Electronics                 & 395           & 398           & 400           & 409           & 399           \\
Gift Cards                  & 140           & 122           & 146           & 144           & 147           \\
Handmade Products           & 385           & 397           & 405           & 412           & 400           \\
Industrial And Scientific   & 404           & 404           & 384           & 392           & 368           \\
Musical Instruments         & 398           & 385           & 403           & 393           & 384           \\
Toys And Games              & 410           & 402           & 378           & 402           & 412           \\
\textbf{Total}              & \textbf{3300} & \textbf{3300} & \textbf{3300} & \textbf{3300} & \textbf{3300} \\ \bottomrule
\end{tabular}
\label{tab:review-dist}
\end{table}

The authors of MAGE reported that their AIG Yelp reviews were noticeably more positive: around 40\% of AIG reviews were classified as positive, while only 30\% of human reviews were positive. To account for this potential discriminator, we balanced review generation by star rating (rather than by product category), as 1-star ratings tend to be more negative than positive. 

For human reviews, we sampled 600 reviews for each star rating and product category combination: 400 reviews for training and 100 reviews each for validation and testing. We sampled from all 34 review categories for a total of 102,000 human reviews.

\subsubsection{Abstracts}

For abstracts, we also use one-shot prompting, prompting the LLM to generate the abstract given an actual (human-written) paper's title. For the one-shot example, we provided the LLM with a randomly selected abstract within the same subject category. Each LLM generated 3,000 training abstracts and 1,000 validation and testing abstracts. 

We generated abstracts for the 20 most frequent individual categories (i.e., no papers cross-listed in multiple categories), including astronomy, mathematics, computer science, and physics. From those same 20 categories, we randomly sampled 80,000 human abstracts for training and 10,000 for validation and testing each (5,000 human abstracts per category across all splits). 

\subsubsection{News}

Each LLM generated 1,600 texts (960 training texts, and 320 texts each for validation and testing). To generate news articles, in each LLM's system prompt, we instructed it to act like a journalist from one of eight news outlets in the US and UK and to copy the style of a given human-written article $a$. We then provide another human-written article $b$'s title and the first 20 tokens from $b$'s content (using GPT-4o-mini's tokenizer from the \verb|tiktoken| library). We did this to provide additional context about the article we want to generate. Each LLM generated 200 articles for each news outlet. 

The selected eight news outlets were:
\begin{itemize}
    \item The Times (UK)
    \item The Sunday Times (UK)
    \item The Guardian (UK)
    \item The Daily Telegraph (UK)
    \item Fox News (US)
    \item CNN (US)
    \item ABC (US)
    \item NBC (US)
\end{itemize}

We selected the UK newspapers based on their popularity across all adults in a survey conducted by YouGov (based on 2025 first-quarter rankings) \citep{yougovMostPopular}. For the US news outlets, the Pew Research Center found that these four sources were the most popular among US adults for political information (survey conducted in September 2024) \citep{pewresearchAmericansSources}. 

We selected these two sets of news outlets due to slight differences in UK and US English --- LLMs' English usage tends to skew more towards Standard American English (SAE), so we suspect that classifiers might be more exposed to SAE than other English dialects, leading to a bias in detection \citep{lee2025trans}.

As an example, when we prompted Gemini 1.5 Flash to write a news article as a journalist from The Times (UK), it inserted the alternative spelling of ``analyze'' in a sentence:
\begin{blockquote}
    Online forums are buzzing with amateur \textbf{analyses}, pixel-by-pixel examinations, and debates about the photographic evidence.
\end{blockquote}
\begin{table}[h!]
\centering
\caption{Distribution of subject matter across AIG news articles.}
\begin{tabular}{@{}ccc@{}}
\toprule
\textbf{real} & \textbf{subject} & \textbf{total} \\ \midrule
fake          & US News          & 1120           \\
fake          & Politics         & 1440           \\
real          & World News        & 2160           \\
fake          & News             & 2200           \\
real          & Politics News     & 2240           \\
fake          & Government News  & 2568           \\
fake          & Left News        & 2592           \\
fake          & Middle East    & 3280           \\ \midrule
\textbf{}     & \textbf{Total}   & \textbf{17600} \\ \bottomrule
\end{tabular}
\label{tab:news-articles-aig}
\end{table}

We balanced the distribution across news outlets in the system prompts for each subject/truthfulness combination. We use all ISOT human articles and split the dataset into 35,916 training articles, 4,489 validation articles, and 4,493 testing articles. 

\subsubsection{Student Essays}

\subsubsubsection{Ivy Panda}

We randomly sampled 100,000 essays from the dataset and allocated 80,000 for training and the remaining 20,000 for validation and testing. However, these essays do not have prompts themselves. Thus, we employed a mirror prompting strategy similar to Pangram Labs \citep{emi-etal-2025-pangram}. Using GPT-4o-mini, we provided the LLM with the human essay and asked it to generate the prompt that could have inspired this essay. We generated 5,500 training prompts and 3,300 prompts each for validation and testing.
\begin{figure}[h!]
    \centering
    \includegraphics[width=1.0\linewidth]{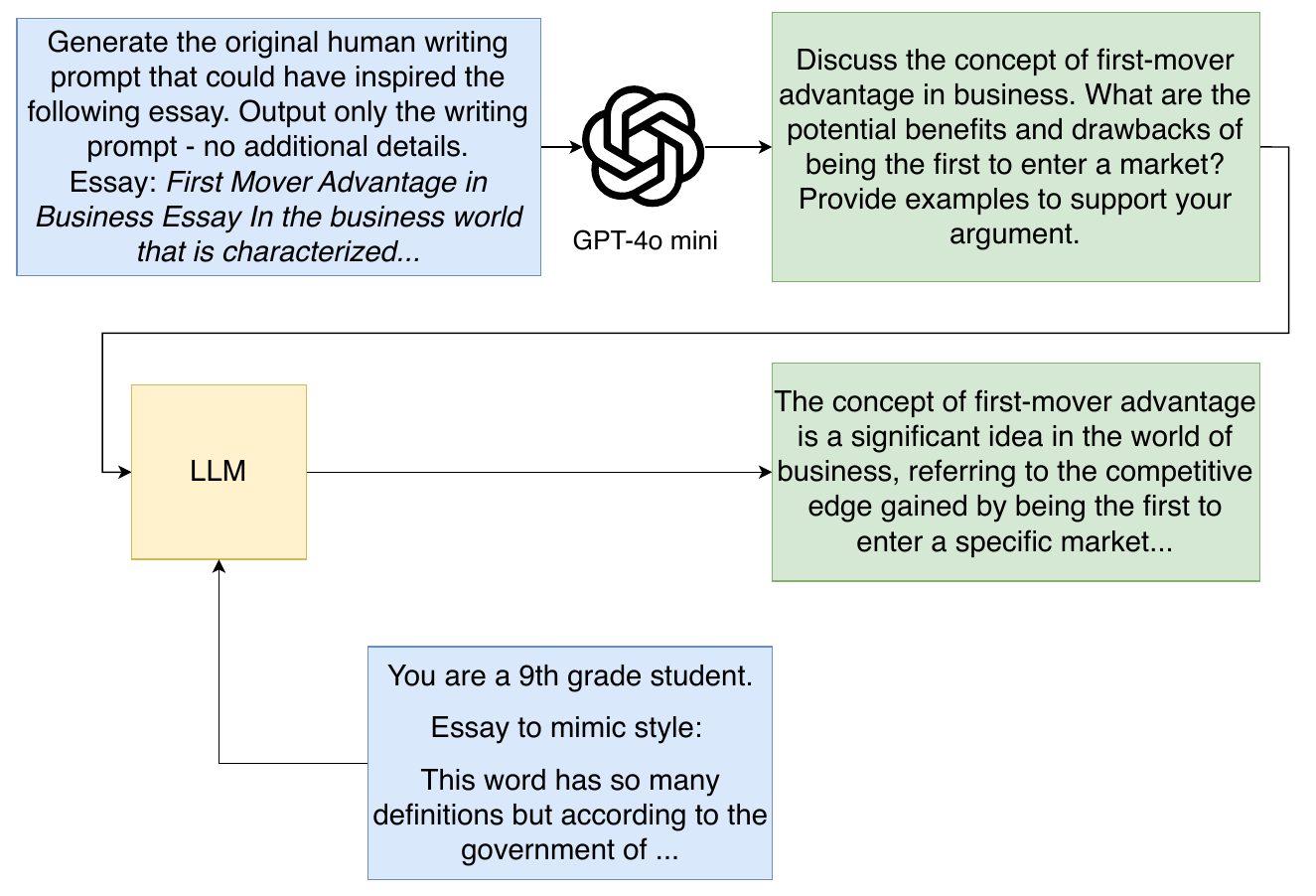}
    \caption{Mirror prompting strategy used to generate essay prompts.}
    \label{fig:mirror-prompt}
\end{figure}
During the essay generation process, we prompted each LLM with a mirror prompt and gave it a random essay to mimic the style. Additionally, in the system prompt, we instructed the LLM to act as either a 9th, 10th, 11th, or 12th-grade level student or an undergraduate student. This is to mitigate the chances of an LLM outputting an essay that is ``too good to be true'' \citep{scarfe2024real}. In the University of Reading study, markers were informed to be wary of submissions that appeared to be above a typical undergraduate level. We randomly selected a grade level from the following distribution: 50\% chance of an undergraduate, and a 12.5\% chance of selecting any grade level from 9th to 12th grade.

\subsubsubsection{ELLIPSE}
 Given its relatively small size, we use the entire corpus. 3,128 texts are selected for training, 783 for validation, and 2,571 for testing. The ELLIPSE corpus contains additional data for each essay, including the prompt and grade level. We use these prompts and grade levels to prompt the LLMs, although we randomly sample from the respective distributions in the corpus rather than uniformly sampling. We use a similar prompt structure to the Ivy Panda subset, but this time we explicitly tell the LLM to act like an ELL (English Language Learner) student. Each LLM generated 220 training texts and 88 validation and testing texts for 4,356 AIG texts.

\subsubsection{Creative Writing}

 Using one-shot prompting, we provide the LLM with a randomly selected story $s$ as an example to base the style on (only the first 300 tokens are given). Then, we provide the LLM with the $s$'s original writing prompt to generate for. Each LLM generates 500 training texts and 200 texts for validation and testing. For human prompts, we sample 50,000 texts (along with their prompts) for training and 10,000 each for validation and testing. 

\subsection{Adversarial Texts}

Our adversarial texts are texts from continued pre-trained (CPT) Llama 3.2 1B (Instruct) models. Creating these texts required two steps: further pre-training the 1B model and prompting the fine-tuned model. 

We selected the Llama 3.2 1B Instruct model due to its popularity and small size relative to other LLMs. This makes it an ideal candidate for potential adversaries to train or fine-tune fully. Also, SLMs (Small Language Models) have not been explored as deeply in terms of AIG text detection as their LLM counterparts. 

\subsubsection{Continued Pre-training}
We continue pre-training 1B models using human texts from a selected domain. We performed quality checks on texts using the TextDescriptives library \citep{Hansen2023}. The quality checks are designed using metrics researchers at Google use to pre-train language models \citep{rae2021scaling} \citep{raffel2020exploring}. There are two types of metrics: heuristic (e.g., number of stop words, out-of-vocabulary ratio, etc.) and repetitious (e.g., duplicate lines, duplicate n-gram character fraction, etc.). We used human texts that passed both sets of metrics for CPT. We report the final numbers of filtered human texts by split and domain in Table \ref{tab:fine-tuned-corpora-size}. 

\begin{table}[h!]

\centering
\caption{CPT corpora size across domains and splits.}
\begin{tabular}{@{}llll@{}}
\toprule
\multicolumn{1}{c}{\textbf{Domain}} & \multicolumn{1}{c}{\textbf{Training}} & \multicolumn{1}{c}{\textbf{Validation}} & \multicolumn{1}{c}{\textbf{Testing}} \\ \midrule
Tweets                              & 27666                                 & 3360                                    & 3411                               \\
Reviews                             & 55290                                 & 13765                                   & 13798                              \\
Abstracts                           & 74762                                 & 9350                                    & 9373                               \\
News                                & 33960                                 & 4243                                    & 4258                               \\
Essays                              & 79882                                 & 10297                                   & 11714                              \\
Creative Writing                    & 43361                                 & 8697                                    & 8713                               \\ \bottomrule
\end{tabular}
\label{tab:fine-tuned-corpora-size}
\end{table}

To prevent data leakage across splits, we pre-train 18 models, one for each split and domain combination. This division also helps simulate real-world conditions better: it is unlikely that the attacker's and defender's pre-training corpora will be an exact match. We perform full-parameter pre-training using the APOLLO-Mini optimizer variant. We train all eighteen models using identical parameters, except for the maximum length of each input text to be used. For tweets, we used 128 tokens; reviews and abstracts had 256 tokens; the remaining three domains had 512 tokens. We pre-train using Kaggle's two T4 GPUs for models from the first three domains. We used one A100 GPU from Cambridge's High Performance Computing (HPC) service for the last three domains' models. 

\subsubsection{Generation}

To account for differences in text length between human and AIG texts, we observed the distribution of text lengths (by counting Llama 3.2 1B tokens) in each domain and split combination for all human texts. Then, we took all possible text lengths between the 25th and 75th percentiles (inclusively) to exclude possible outliers. We then randomly sampled from this clipped distribution, $C$, to get an individual text length $T$ to ``generate'' for a text. 

We use a continuation prompt, in which the fine-tuned model is given the first $H$ tokens from a human text and is expected to generate up to $N$ tokens. Given $T$, where $T \sim C$, we set $H = \lfloor T/3 \rfloor$ and $N = T - H$. The final text includes the $H$ tokens from the human text, followed by $N$ tokens. \cite{zhang-etal-2024-llm} refers to this type of text as \textit{mixcase} --- text that contains both human and LLM text. They also demonstrate that mixcase texts pose a challenge for existing detectors.

We selected a temperature of 1.1 and a top-$p$ value of 1 during the generation process. Additionally, we set the top-$k$ parameter to 100. This parameter restricts the \textit{number} of candidate tokens during generation. We use these parameters across all fine-tuned models, regardless of split and domain. We justify this decision as during our initial detection experiments against the MAGE classifier, we observed that this particular set of generation parameters yielded low detection scores. It is fair to assume that both an attacker and a defender may arrive at similar results when scoping out weaknesses in existing detectors. 

To isolate the effects of fine-tuning against the mixcase effect, we compare CPT generations to the original Llama 3.2 1B Instruct generations. We use the same prompts as well as identical generation parameters.

\newpage
\subsection{Final Dataset Statistics}

We report the non-adversarial and adversarial statistics in Tables \ref{tab:dactyl-nonadversarial} and \ref{tab:adversarial-dist}. We do not generate training and validation texts using the non-trained 1B Instruct model, as it serves as a comparison against our CPT models. 
\begin{table}[h!]
\centering
\caption{Non-adversarial distribution of DACTYL texts by split and domain.}
\begin{tabular}{@{}cccccccc@{}}
\toprule
                                     & \multicolumn{2}{c}{Training}        & \multicolumn{2}{c}{Validation}     & \multicolumn{2}{c}{Testing}        & Total  \\ \midrule
Domain                               & Human  & AI                         & Human & AI                         & Human & AI                         &        \\ \midrule
\multicolumn{1}{c|}{Tweets}          & 56801  & \multicolumn{1}{c|}{16500} & 7066  & \multicolumn{1}{c|}{6600}  & 7080  & \multicolumn{1}{c|}{6600}  & 100647 \\
\multicolumn{1}{c|}{Reviews}         & 68000  & \multicolumn{1}{c|}{11000} & 17000 & \multicolumn{1}{c|}{2750}  & 17000 & \multicolumn{1}{c|}{2750}  & 118500 \\
\multicolumn{1}{c|}{Abstracts}       & 80000  & \multicolumn{1}{c|}{33000} & 10000 & \multicolumn{1}{c|}{11000} & 11000 & \multicolumn{1}{c|}{11000} & 155000 \\
\multicolumn{1}{c|}{News}            & 35916  & \multicolumn{1}{c|}{10560} & 4489  & \multicolumn{1}{c|}{3520}  & 4493  & \multicolumn{1}{c|}{3520}  & 62498  \\
\multicolumn{1}{c|}{Student Essays}  & 83128  & \multicolumn{1}{c|}{7920}  & 10783 & \multicolumn{1}{c|}{4268}  & 12571 & \multicolumn{1}{c|}{4268}  & 122938 \\
\multicolumn{1}{c|}{Writing Prompts} & 50000  & \multicolumn{1}{c|}{5500}  & 10000 & \multicolumn{1}{c|}{2200}  & 10000 & \multicolumn{1}{c|}{2200}  & 79900  \\ \midrule
Total                                & 373845 & 84480                      & 59338 & 30338                      & 61144 & 30338                      & -      \\ \midrule
Total                                & \multicolumn{2}{c}{458325}          & \multicolumn{2}{c}{89676}          & \multicolumn{2}{c}{91482}          & 639483 \\ \bottomrule
\end{tabular}
\label{tab:dactyl-nonadversarial}
\end{table}

\begin{table}[h!]
\centering
\caption{Distribution of 1B-Instruct generations by domain and split. Base refers to a non-CPT model.}
\begin{tabular}{@{}cccccccc@{}}
\toprule
                                      & \multicolumn{2}{c}{Training}     & \multicolumn{2}{c}{Validation}   & \multicolumn{2}{c}{Testing}      &       \\ \midrule
Domain                                & Base & CPT            & Base & CPT             & Base & CPT            & Total \\ \midrule
\multicolumn{1}{c|}{Tweets}           & 0    & \multicolumn{1}{c|}{1500} & 0    & \multicolumn{1}{c|}{600}  & 600  & \multicolumn{1}{c|}{600}  & 3300  \\
\multicolumn{1}{c|}{Reviews}          & 0    & \multicolumn{1}{c|}{1000} & 0    & \multicolumn{1}{c|}{250}  & 250  & \multicolumn{1}{c|}{250}  & 1750  \\
\multicolumn{1}{c|}{Abstracts}        & 0    & \multicolumn{1}{c|}{3000} & 0    & \multicolumn{1}{c|}{1000} & 1000 & \multicolumn{1}{c|}{1000} & 6000  \\
\multicolumn{1}{c|}{News}             & 0    & \multicolumn{1}{c|}{960}  & 0    & \multicolumn{1}{c|}{320}  & 320  & \multicolumn{1}{c|}{320}  & 1920  \\
\multicolumn{1}{c|}{Student Essays}   & 0    & \multicolumn{1}{c|}{720}  & 0    & \multicolumn{1}{c|}{388}  & 388  & \multicolumn{1}{c|}{388}  & 1884  \\
\multicolumn{1}{c|}{Creative Writing} & 0    & \multicolumn{1}{c|}{500}  & 0    & \multicolumn{1}{c|}{200}  & 200  & \multicolumn{1}{c|}{200}  & 1100  \\ \midrule
Total                                 & 0    & 7680                      & 0    & 2758                      & 2758 & 2758                      & 15954 \\ \bottomrule
\end{tabular}
\label{tab:adversarial-dist}
\end{table}

\section{Experiments}
\subsection{Classifiers Used}
We evaluate both pre-trained classifiers (i.e., existing AIG text detectors) and our DACTYL-trained classifiers on the DACTYL test set. 
\ subsubsection{Pre-trained Classifiers}
\begin{table}[h!]
\centering
\caption{List of pre-trained classifiers evaluated on DACTYL.}
\begin{tabularx}{\textwidth}{@{}cccX@{}}
\toprule
Classifier    & Type             & Parameter Count & Training Set                             \\ \midrule
Binoculars    & Zero-Shot        & 14B            & N/A                                      \\
DAIGT         & DeBERTa-V3-Large & 434M            & SlimPajama,Persuade Corpus, DAIGT        \\
Desklib v1.01 & DeBERTa-V3-Large & 434M            & RAID                                     \\
e5-LoRA       & e5-small         & 33.4M           & Subset of RAID, AIG tweets               \\
Fakespot      & RoBERTa-base     & 125M            & 14 AIG datasets, including RAID          \\
MAGE          & Longformer       & 149M            & MAGE                                     \\
Pangram       & Transformer      & Unknown         & Unknown, commercial grade                \\
AIContentDetector (AICD)     & RoBERTa-base     & 125M            & Unknown, commercial?                     \\
SuperAnnotate & RoBERTa-large    & 354M            & RAID, Wikipedia, ELI5, Scientific Papers \\ \bottomrule
\end{tabularx}
\label{tab:classifier-table}
\end{table}
We use a variety of pre-trained classifiers as shown in Table \ref{tab:classifier-table}. We selected Binoculars to evaluate how effective relying on a variant of perplexity would be against the DACTYL test set. Desklib, e5-LoRA, and Pangram showcased strong performance on the RAID test set \citep{huggingfaceDesklibaitextdetectorv101Hugging} \citep{huggingfaceMayZhoue5smallloraaigenerateddetectorHugging} \citep{emi-etal-2025-pangram} \citep{dugan_raid_2024}. Desklib and e5-LoRA were trained on the RAID dataset; Pangram is an enterprise-grade AI text detector, and not all details are available publicly. DAIGT DeBERTa-V3-Large obtained seventh place (out of 4,358 teams) in Kaggle's Detect AI-Generated Text (DAIGT) competition, which focused on student essay detection \citep{Mei2024-zm} \citep{llm-detect-ai-generated-text}. This particular DeBERTa-V3-large model is unique due to its two-stage training: it initially involves training on around 500,000 AIG texts from the SlimPajama corpus, followed by domain-adaptive fine-tuning specifically on AIG essays \citep{Mei2024-zm}. FakeSpot is trained on a massive corpus spanning 14 datasets, including additional supporting data from newer LLMs such as GPT-4o and GPT-4o-mini. Fakespot also includes training on the Amazon 2023 review corpus \citep{githubApollodftdataMain}. We evaluated the MAGE Longformer due to its robustness on unseen domains and models \citep{li_mage_2024}. Similar to Pangram, AICD is a commercial detector available from RapidAPI, but the model is available on HuggingFace\citep{rapidapiAIContentDetector} \citep{huggingfacePirateXXAIContentDetectorHugging}. Unfortunately, the commercial nature of these models means that exact training details are unavailable. SuperAnnotate was selected as its model coverage includes four out of the five LLM families used in developing DACTYL: GPT (OpenAI), Anthropic, Mistral, and Llama \citep{huggingfaceFakespotairobertabaseaitextdetectionv1Hugging}. 

We list Binoculars as having 14 billion parameters as it uses two Falcon 7B models \citep{hans_spotting_2024}.
\ subsubsection{DACTYL Trained classifiers}

We train four transformer models on the DACTYL training split: BERT-tiny, TinyBERT, distilRoBERTa-base, and ModernBERT-base. We train each classifier for five epochs (as used by the MAGE classifier), only keeping the model with the best validation tpAUC \citep{li_mage_2024}. We use a learning rate of 2e-5 as suggested by \cite{sun2019fine}. To speed up training, we use a larger batch size of 64. We use PyTorch's \verb|BCELossWithLogits| and the AdamW optimizer for our initial loss function and optimizer. \verb|BCELossWithLogits| is numerically more stable than traditional cross-entropy loss followed by a sigmoid computation (done to rescale logits from 0 to 1) \citep{pytorchBCEWithLogitsLossx2014}. The AdamW optimizer's decoupled weight decay increases generalization performance over the standard Adam optimizer \citep{loshchilov2017decoupled}. 

For the DeBERTa-V3-large, we trained with the same parameters as the other four models but only for one epoch and a batch size of 16 due to GPU memory limitations, as we observed a steep drop-off in tpAUC in the second epoch (validation tpAUC went from 75 to 0). This performance hit suggests that the model overfits rapidly with just one epoch. 

According to the technical documentation for LibAUC, pre-training a classifier with cross-entropy followed by DXO can increase performance \citep{yuan2023libauc}. We take the ``pre-trained'' models (i.e., models trained using \verb|BCELossWithLogits|), and further train those models using the tpAUC loss function. We reset the classification head's parameters as shown in the tutorials. Since this loss function requires each batch to have a positive (AIG) and negative (human) sample, we use a controlled data sampler to ensure each batch has a fixed proportion of positive samples. For the first four models, we set the proportion (sampling rate) equal to 0.5 with a batch size of 64, oversampling the positive class. We reduce the learning rate to 1e-5 --- decaying the learning rate can help the model learn nuances in the training set better \citep{you2019does}. 

For DeBERTa-V3-large, since the model's performance worsens, we do not apply DXO to the cross-entropy pre-trained model; instead, we train DeBERTa-V3-large from scratch (i.e., no cross-entropy training beforehand). The learning rate and sampling rate are the same for the four smaller models, but the batch size is still 16. We note that cross-entropy pre-training is not necessary for DXO, according to the experiments conducted by Yuan et al., but it does improve performance \citep{yuan2023libauc}. 

We refer to binary cross-entropy trained classifiers as BCE classifiers and those trained with tpAUC loss as DXO (deep X-risk optimized) classifiers. 

\subsection{Out-of-Distribution (OOD) Test Set}

For our DACTYL classifiers, we evaluate their generalization capability to existing AIG text detection test sets. This additional evaluation determines if our classifiers are ``overfitting'' to the DACTYL test set. Adversarially trained deep-learning models are prone to overfitting and struggle with unmodified examples \citep{rice2020overfitting}. We select from the following data sources in constructing our OOD test set, with full statistics reported in Table \ref{tab:ood-description}. 

\textbf{AuTexTification.} The AuTexTification test set has 21,832 texts and includes two domains: Amazon reviews and news articles \citep{sarvazyan2023overview}. Given that the DACTYL dataset has these two domains, we can determine if our classifiers can generalize. \cite{sarvazyan2023overview} used the BLOOM family and older OpenAI models to generate the AIG texts. 

\textbf{BEEMO.} The Benchmark of Expert-Edited Machine-generated Outputs (BEEMO) dataset addresses a more recent gap in AIG text detection, focusing on mixcase detection \citep{artemova2024beemo}. AIG texts are edited by humans, GPT-4o, or Llama 3.1 70B. Since BEEMO's texts often blur the line between AI and human, we use \cite{artemova2024beemo}'s binary classification formulation: only \textbf{pure} human texts are considered human; any other text should be labelled as AIG. This definition also means that a human-edited LLM-generated text should be considered AIG. Since there are no clearly defined splits, we use all 19,683 texts. BEEMO evaluates how well our classifiers can detect AIG text surrounded by human text. 

\textbf{MAGE.} Spanning 27 LLMs and seven major domains, MAGE is one of the most exhaustive test sets \citep{li_mage_2024}. We use the entire test set aggregated over all domains for 60,743 texts. This dataset represents an extreme robustness test for our DACTYL classifiers. 

\textbf{DREsS and AIG-ASAP.} Dataset for Rubric-based Essay Scoring (on English as a Foreign Language Writing) is one of the latest student essay datasets, released in 2024 \citep{yoo2024dress}. The DREsS dataset consolidates existing AES (Automated Essay Scoring) human essay datasets. Additionally, it provides a new student essay (written in English) dataset from students in an undisclosed university in South Korea \citep{yoo_dress_2024}. Since DREsS only includes human texts, we include texts from the AIG-ASAP dataset \citep{peng_hidding_2023}. This dataset contains AIG essays that have gone through adversarial perturbations, including paraphrasing, sentence substitution, and synonym substitution. AIG-ASAP uses the GPT-3.5, GPT-4, and Vicuna 7B models to generate their essays. The total size of the DREsS and AIG-ASAP is 9,078 texts. We included these datasets to mimic a possible real-world deployment for our classifiers --- likely, the text distribution of DREsS/AIG-ASAP differs noticeably from the Ivy Panda/ELLIPSE training set.

\begin{table}[h!]
\centering
\caption{OOD test set sources and human/AIG splits.}
\begin{tabular}{@{}cccc@{}}
\toprule
Dataset         & Human & AIG                        & Total  \\ \midrule
AuTexTification & 10642 & \multicolumn{1}{c|}{11190} & 21832  \\
BEEMO           & 2187  & \multicolumn{1}{c|}{17496} & 19683  \\
MAGE            & 30265 & \multicolumn{1}{c|}{30478} & 60743  \\
DREsS+AIG-ASAP  & 3958  & \multicolumn{1}{c|}{5120}  & 9078   \\ \midrule
Total           & 47052 & 64284                      & 111336 \\ \bottomrule
\end{tabular}
\label{tab:ood-description}
\end{table}

\section{Results \& Analysis}

\subsection{Non-Adversarial Results}

\subsubsection{Pre-trained Classifiers Results}
We report the four X-risks for the nine pre-trained classifiers, averaged across the six domains, in Table \ref{tab:pt-classifier-nonadversarial}. We set pAUC($\beta=5\%$) and tpAUC($\alpha=50\%,\beta=5\%$). We use a maximum FPR of 5\% consistent with \cite{dugan_raid_2024}. We set the minimum TPR at 50\% to minimize non-zero tpAUC scores (AICD and MAGE are the only classifiers with tpAUC scores less than 10).

\begin{table}[h!]
\centering
\caption{Mean scores across the six domains for each pre-trained classifier.}
\begin{tabular}{@{}ccccc@{}}
\toprule
Classifier    & tpAUC(50\%, 5\%) & pAUC(5\%)  & AUC   & AP    \\ \midrule
Pangram       & \textbf{76.31} &\textbf{ 93.18} & \textbf{95.99} & \textbf{95.97} \\
Desklib       & 66.09 & 87.73 & 91.03 & 89.56 \\
Fakespot      & 61.12 & 87.39 & 94.45 & 92.24 \\
Binoculars    & 50.72 & 82.95 & 93.0  & 86.71 \\
SuperAnnotate & 43.56 & 78.99 & 89.69 & 83.02 \\
DAIGT         & 24.99 & 74.76 & 81.92 & 77.12 \\
e5-LoRA       & 20.73 & 72.41 & 87.44 & 76.82 \\
MAGE          & 3.56  & 62.18 & 78.88 & 66.17 \\
AICD          & 2.67  & 61.78 & 74.57 & 63.49 \\ \bottomrule
\end{tabular}
\label{tab:pt-classifier-nonadversarial}
\end{table}

We highlight the lack of separation between the top classifiers in terms of AUC: three out of the top four classifiers have an AUC $\geq 93$, but only one, Pangram, has a tpAUC $\geq 70$. The next highest mean tpAUC, Desklib, is around 10 points less. Also, the mean tpAUC (and pAUC) scores are substantially smaller than the mean AUC and AP scores. This discrepancy supports the notion that the AUC and AP scores are ``boosted'' by performance in non-deployable regions (at high FPR regions). 

\begin{figure}[h!]
    \centering
    \includegraphics[width=1.0\linewidth]{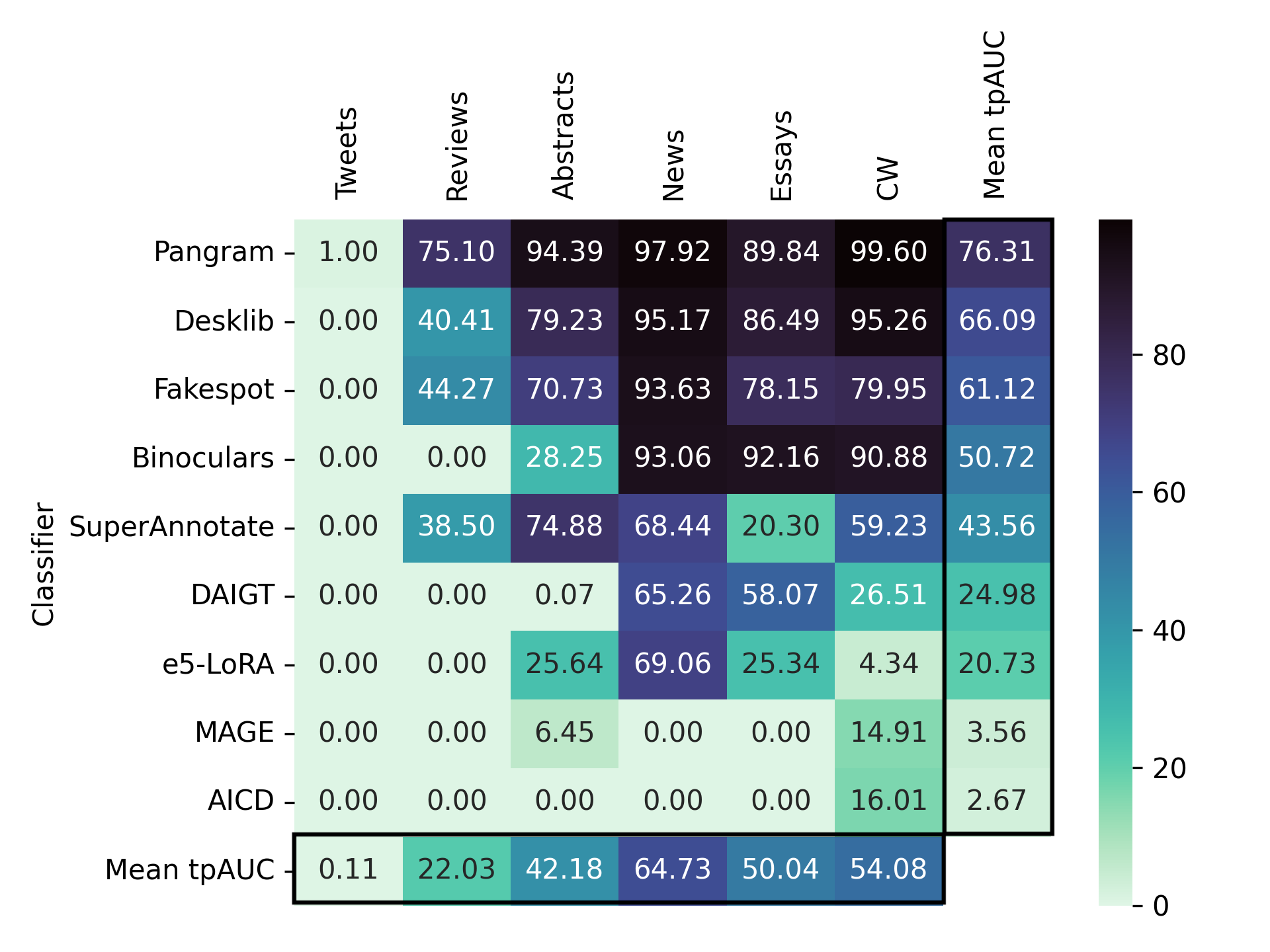}
    \caption{tpAUC scores by domain and classifier.}
    \label{fig:baseline-non-adversarial-domain-heatmap}
\end{figure}

Not surprisingly, the classifiers struggled significantly on tweets with only Pangram having a non-zero tpAUC as shown in Figure \ref{fig:baseline-non-adversarial-domain-heatmap}. In contrast, the news category was considerably easier, with a mean tpAUC of 64.73, 10 points higher than the next easiest domain (creative writing, CW).  

The zero-shot Binoculars method is quite capable of outperforming various transformer-based classifiers. Binoculars is the only other classifier that outscored Pangram in one domain (student essays). However, this method struggles with shorter text domains: the tpAUC score is zero for tweets and reviews and less than 30 for abstracts. These results indicate that perplexity (and cross-perplexity) are unreliable signals on shorter texts. 

\begin{figure} [h!]
    \centering

    \includegraphics[width=1.0\linewidth]{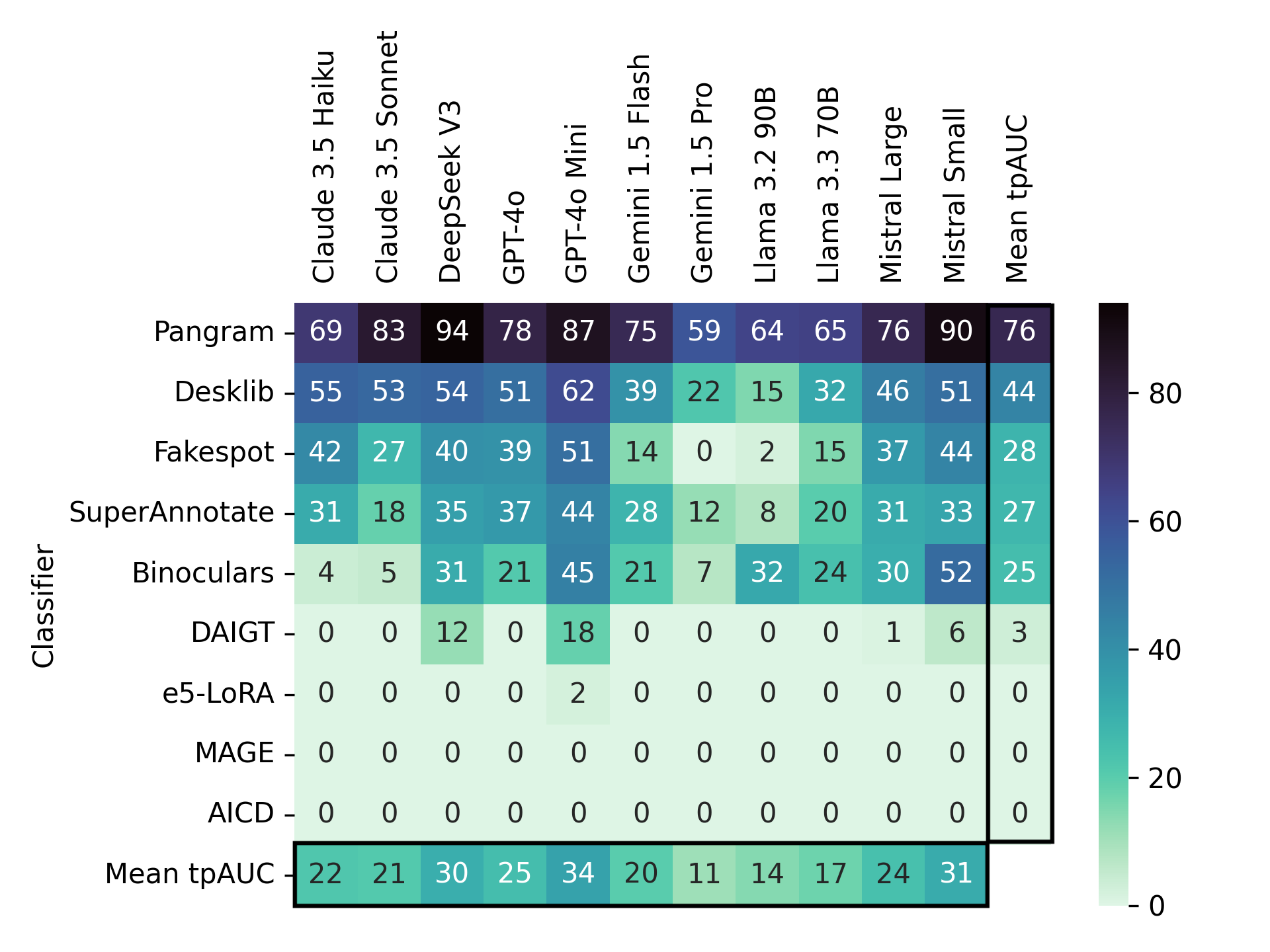}
        \caption{tpAUC(50\%, 5\%) scores by model and classifier.  }
    \label{fig:baseline-non-adversarial-model-heatmap}
\end{figure}

Similarly to \cite{thorat_which_2024}, we investigate if detection performance varies across LLMs, as shown in Figure \ref{fig:baseline-non-adversarial-model-heatmap}. The Gemini 1.5 and Llama families are the most challenging to detect: the highest mean tpAUC for any of these four models is 20 (Gemini 1.5 Flash). Five of the nine classifiers have a tpAUC score of 0 on the Gemini 1.5 Pro model. Detectors may struggle on Gemini 1.5 models due to their Mixture of Experts (MoE) architecture --- multiple smaller neural networks function as a unified network \citep{blogNextgenerationModel}. Gemini 1.5 might be activating smaller networks (rather than its entire network) for specific text generation tasks which could degrade detection. Llama 3.2's difficulty might stem from its architecture needing to support vision tasks \citep{metaLlama32}. 

We also observe that larger LLMs tend to outscore their smaller counterparts within the same family: Claude 3.5 Sonnet, GPT-4o, Gemini 1.5 Pro, Llama 3.2 90B, and Mistral Large all have smaller mean tpAUC scores than Haiku, GPT-4o Mini, Flash, Llama 3.3 70B, and Mistral Small models, respectively. Using Binoculars performance as a proxy for perplexity, the larger LLMs typically seem to generate more unpredictable (and diverse) texts. 

Curiously, despite being one of the newest (and largest) models launched, DeepSeek-V3 remains noticeably easy to detect: Pangram achieves the highest tpAUC score of any (classifier, LLM) combination on DeepSeek-V3, with a tpAUC score of 94. The Llama family, where each model has less than 100 billion parameters, outperforms DeepSeek-V3 by a considerable margin. DeepSeek-V3 is also the third-easiest model to detect out of the 11 LLMs, only surpassed by GPT-4o mini (34) and Mistral Small (31). We highlight this due to DeepSeek-V3's massive size of 671 billion parameters, which suggests that model size may not be the \textit{sole} determining factor in determining which models are challenging to distinguish. This behavior of DeepSeek-V3 being ``easy to detect'' was confirmed by another commercial detector, Originality AI. They point out that newer LLMs tend to reduce their classifier accuracy upon release, but did not observe this trend with DeepSeek-V3 \citep{originalityDeepSeekCopy}. \cite{originalityDeepSeekCopy} suspects DeepSeek-V3 may have been trained on other LLMs, such as the OpenAI family. This training process might explain why several classifiers, such as Desklib (released in October 2024), have relatively high tpAUC scores, despite not being exposed to DeepSeek-V3 generations in their training process. 

\cite{thorat_which_2024} demonstrated that the OpenAI family posed a greater challenge to detectors. However, the pre-trained classifiers tested did not struggle significantly against OpenAI LLMs. It is likely that, given the popularity and accessibility of GPT-4o and GPT-4o mini, dataset developers use OpenAI-generated texts extensively, which means that AIG text detectors are exposed more to those texts.

\subsubsection{Binary Cross-Entropy vs tpAUC Loss}
\label{sec:cross-entropy-vs-tpauc-loss}
\begin{figure}[h!]
    \centering
    \includegraphics[width=1\linewidth]{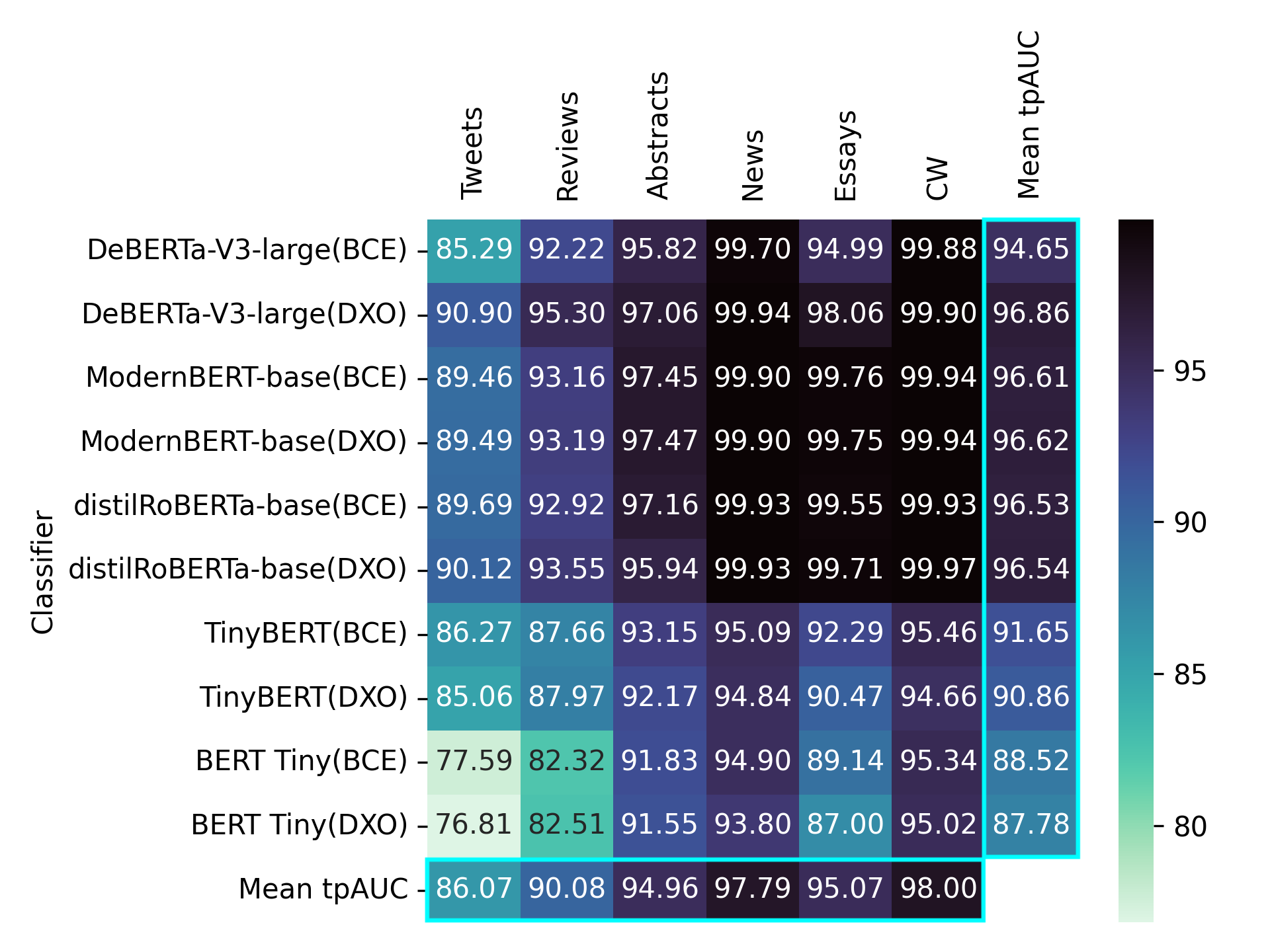}
    \caption{tpAUC(50\%, 5\%) by domain and classifier. BCE indicates a classifier trained with binary cross-entropy loss. DXO indicates a deep X-risk (in this case, tpAUC loss) optimized classifier. }
    \label{fig:dacty-trained-non-adv}
\end{figure}

For the DACTYL-trained classifiers, we observe fine-tuning with the tpAUC loss function makes more noticeable improvements for larger models in Figure \ref{fig:dacty-trained-non-adv}. BERT-tiny and TinyBERT had worse tpAUC(50\%,5\%) after training with the tpAUC loss function, while distilRoBERTa and ModernBERT-base had negligible gains. In contrast, DeBERTa-V3-large had a noticeable increase of more than 2 points.

Similar to the pre-trained classifiers, DACTYL-trained classifiers struggled more on the shorter text domains such as tweets, reviews, and abstracts. The smaller classifiers, such as BERT-Tiny, had a harder time on these domains. However, this is significantly higher than the pre-trained classifiers. 

As model size grows, the tpAUC score grows more slowly: you can obtain a tpAUC score of 96.62 with a distilRoBERTa-base (82 million parameters), but only a tpAUC score of 96.86 with a DeBERTa-V3-large (434 million parameters). This slowdown in the tpAUC score increase suggests the model's capacity to learn the training data is being reached. We can confirm results by checking the AUC scores in Table \ref{tab:dacty-trained-non-adv-auc}. 

\begin{table}[h!]
\centering
\caption{Mean AUC scores by domain and loss function used to train classifiers on the non-adversarial DACTYL test set. Bold values indicate maximum values.}
\begin{tabular}{@{}ccc@{}}
\toprule
                   & \multicolumn{2}{c}{AUC}      \\ \midrule
Classifier         & BCEWithLogitsLoss & tpAUC Loss \\ \midrule
BERT-tiny          & \textbf{99.13} & 99.04 \\
TinyBERT           & \textbf{99.35} & 99.23 \\
distilRoBERTa-base & 99.72 & \textbf{99.73} \\
ModernBERT-base    & 99.75 & \textbf{99.75} \\
DeBERTa-V3-large   & 99.6  & \textbf{99.78} \\\bottomrule
\end{tabular}
\label{tab:dacty-trained-non-adv-auc}
\end{table}

The AUC scores are greater than $\geq 99.0$, indicating that all models can easily distinguish between human and AIG texts on the DACTYL test set (although their performance in the upper left region of the ROC curve varies). 

\begin{figure}[h!]
    \centering
    \includegraphics[width=1.0\linewidth]{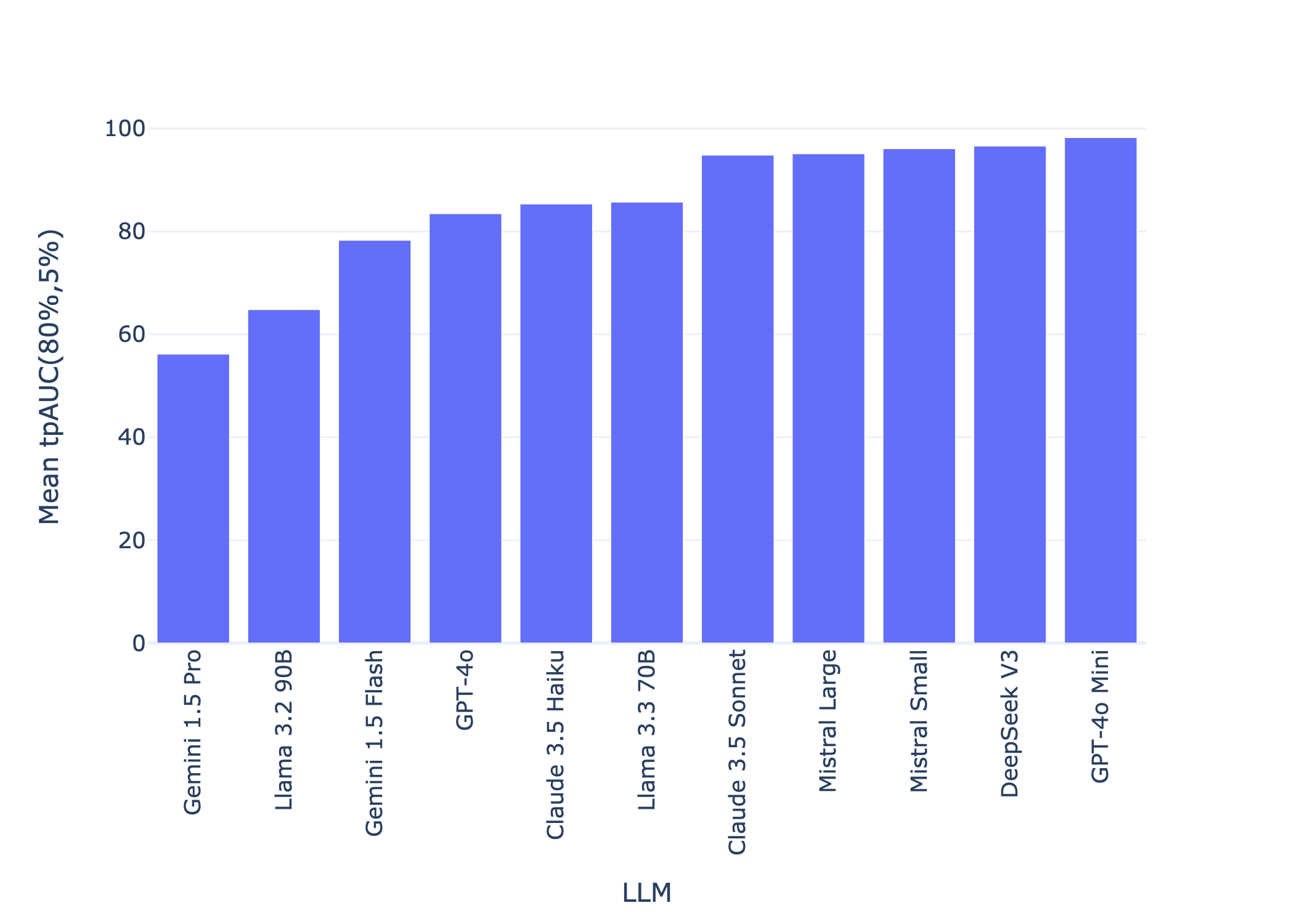}
    \caption{Mean tpAUC(80\%, 5\%) score of all 10 DACTYL trained classifiers by model.}
    \label{fig:dactyl-trained-non-adversarial-model-model}
\end{figure}

We evaluate LLM difficulty as we did with the pre-trained classifiers across all ten DACTYL-trained classifiers in Figure \ref{fig:dactyl-trained-non-adversarial-model-model}; since the AUC scores and tpAUC scores are high, we increase $\alpha$ from 50\% to 80\% to highlight classifier differences better. However, this means that the tpAUC scores are not directly comparable to the pre-trained classifiers' scores due to different $\alpha$ values. We can still compare LLMs' difficulty ``rankings'' between the two sets of classifiers. LLM difficulty somewhat agrees with the pre-trained classifiers: GPT-4o mini, DeepSeek-V3, and Mistral Small are the easiest LLMs, while Gemini 1.5 Pro and Llama 3.2 90B are the most difficult for the DACTYL-trained classifiers. However, the trend of better/larger models being more difficult to detect within the same family does not hold for one case: Claude 3.5 Haiku (mean tpAUC of 85.34) is more difficult to detect than Claude 3.5 Sonnet (94.84). This discrepancy confirms that having more parameters does not guarantee a more ``human'' text. The negligible gap between Mistral Large (95.11) and Small (96.08) supports this somewhat.

\subsubsection{Biases by Domain and Category}
This section looks at possible contributing factors within certain domains (e.g., for reviews, star ratings, for student essays, Ivy Panda vs ELLIPSE performance, etc.). 

We evaluate the pre-trained classifiers by comparing mean tpAUC(50\%, 5\%) scores across each factor. We look at the maximum tpAUC(80\%, 5\%) score per factor for our DACTYL-trained classifiers. We use the maximum scores, rather than the mean, for the trained classifiers as these classifiers have been exposed to these factors already; if, after training, the maximum tpAUC score is still ``low'' across all classifiers, it indicates that particular factor is inherently problematic.  We raise $\alpha$ from 50\% to 80\% as explained in section \ref{sec:cross-entropy-vs-tpauc-loss}. 

We examine four domains: reviews, abstracts, news, and student essays. We exclude tweets because the pre-trained classifiers performed poorly on them. Creative writing is the only major domain without any notable factors within it. 

\subsubsubsection{Reviews}

As suggested by \cite{li_mage_2024}, AIG reviews are more likely to be positive. Intuitively, we expect one-star AIG reviews to be harder to detect than higher-star AIG reviews. 

We investigate if this potential discrepancy influences detection results by stratifying the test set by star rating. Since four pre-trained classifiers had a tpAUC(50\%, 5\%) of 0 across all star ratings, we only focus on pre-trained classifiers that had at least one non-zero tpAUC score, as shown in Table \ref{tab:review-star-pretrained}. 

\begin{table}[h!]
\centering
\caption{tpAUC(50\%, 5\%) by review star for pre-trained classifiers. Bold values indicate the highest scores for a star rating.}
\begin{tabular}{@{}cccccc@{}}
\toprule
              & \multicolumn{5}{c}{Star Rating}       \\ \midrule
Classifier    & 1     & 2     & 3     & 4     & 5     \\ \midrule
DAIGT         & 0.00  & 0.00  & 0.00  & 0.00  & 1.79  \\
Desklib       & 23.34 & 40.30 & 49.83 & 51.24 & 41.74 \\
Fakespot      & 27.87 & 45.75 & 46.71 & 50.58 & 57.60 \\
Pangram       & \textbf{61.79} & \textbf{72.54} & \textbf{75.86} & \textbf{81.87} & \textbf{83.56} \\
SuperAnnotate & 43.79 & 53.00 & 54.78 & 30.49 & 23.70 \\ \midrule
Mean          & 31.36 & 42.32 & 45.44 & 42.84 & 41.68 \\ \bottomrule
\end{tabular}
\label{tab:review-star-pretrained}
\end{table}

Two classifiers --- Pangram and Fakespot --- saw their tpAUC scores increase as star rating increases; Desklib follows this trend until four-star reviews and drops off for five-star reviews. Pre-trained classifiers struggled with one-star reviews significantly compared to other ratings. According to Table \ref{tab:review-star-trained}, our DACTYL-trained classifiers also exhibited this behavior, as the highest tpAUC(80\%, 5\%) was 80.55 for one-star reviews while all other star ratings had scores $\geq 85$.

\begin{table}[h!]
\centering
\caption{Highest tpAUC scores by star rating for the DACTYL-trained classifiers. }
\begin{tabular}{@{}ccc@{}}
\toprule
Stars & max tpAUC(80\%,5\%) & Obtained by   \\ \midrule
1              & 80.55                        & DeBERTa-V3-large(DXO) \\
2              & 86.75                        & DeBERTa-V3-large(DXO) \\
5              & 88.63                        & DeBERTa-V3-large(DXO) \\
4              & 90.38                        & DeBERTa-V3-large(DXO) \\
3              & 94.32                        & DeBERTa-V3-large(DXO) \\ \bottomrule
\end{tabular}
\label{tab:review-star-trained}
\end{table}

This performance drop for one-star reviews highlights an exploitable weakness by adversaries. Independent sellers have already weaponized fake (human) one-star reviews during the COVID-19 pandemic to damage rivals' reputations \citep{bbcAmazonsMurky}. The fact that detectors struggle with these fake negative reviews may amplify the problem in online marketplaces, hurting sellers (and indirectly, consumers). 

\subsubsubsection{Abstracts}
Table \ref{tab:arxiv-pretrained-categories} has the top five easiest and hardest subjects by mean tpAUC for the pre-trained classifiers. 

\begin{table}[h!]
\centering
\caption{Top five most challenging arXiv subjects and top five easiest arXiv subjects by tpAUC for pre-trained classifiers.}
\begin{tabular}{@{}lcc@{}}
\toprule
Ranking & Subject                                & Mean tpAUC(50\%, 5\%) \\ \midrule
1       & Number Theory                          & 32.19                 \\
2       & Combinatorics                          & 35.58                 \\
3       & Algebraic Geometry                     & 36.40                 \\
4       & High Energy Physics - Phenomenology    & 36.81                 \\
5       & Probability                            & 36.96                 \\ \midrule 
16      & Solar and Stellar Astrophysics         & 
57.75                 \\
17      & High Energy Astrophysical Phenomena    & 58.03                 \\
18      & Cosmology and Nongalactic Astrophysics & 58.56                 \\
19      & Computer Vision                        & 59.02                 \\
20      & Astrophysics of Galaxies               & 61.23                 \\ \bottomrule
\end{tabular}
\label{tab:arxiv-pretrained-categories}
\end{table}

Four out of the five most challenging categories are within the mathematics field. Conversely, AIG astrophysics abstracts are generally easier to distinguish from their human-written counterparts. Mathematical abstracts may likely have \LaTeX\ commands in their text, which can degrade detection, as most AIG texts would not have these commands.  

DACTYL-trained classifiers do not appear to struggle as much on mathematical abstracts, as the only math subject that placed in the top five was number theory (Table \ref{tab:arxiv-trained-categories}). Astrophysics abstracts remain easy to detect, with the bottom four subjects belonging to that category.  
\begin{table}[h!]
\caption{Maximum tpAUC scores for challenging and easiest abstracts by subject matter for DACTYL-trained classifiers.}
\small
\centering
\begin{tabular}{@{}cccc@{}}
\toprule
Rank & Subject                                  & tpAUC(80\%, 5\%) & Obtained by              \\ \midrule
1    & Number Theory                            & 90.68            & ModernBERT-base(DXO)    \\
2    & General Relativity and Quantum Cosmology & 92.11            & distilRoBERTa-base(BCE) \\
3    & Analysis of PDEs                         & 92.27            & DeBERTa-V3-large(DXO)   \\
4    & High Energy Physics - Theory             & 92.39            & distilRoBERTa-base(DXO) \\
5    & High Energy Physics - Phenomenology      & 92.77            & distilRoBERTa-base(DXO) \\ \midrule
16   & Mesoscale and Nanoscale Physics          & 97.56            & ModernBERT-base(DXO)    \\
17   & High Energy Astrophysical Phenomena      & 98.00            & distilRoBERTa-base(DXO) \\
18   & Cosmology and Nongalactic Astrophysics   & 98.15            & ModernBERT-base(DXO)    \\
19   & Solar and Stellar Astrophysics           & 98.67            & DeBERTa-V3-large(DXO)   \\
20   & Astrophysics of Galaxies                 & 99.89            & distilRoBERTa-base(DXO) \\ \bottomrule
\end{tabular}
\label{tab:arxiv-trained-categories}
\end{table}

Note that DXO classifiers achieve the highest tpAUC scores in all but one category.

\subsubsubsection{News}

We did not find any significant \textit{consistent} difference in detection performance between ``real'' or ``fake'' news, as shown in Table \ref{tab:news-pretrained-real} among pre-trained classifiers. Four of the nine classifiers struggled with fake AIG news, while the remaining five had the opposite behavior.

\begin{table}[h!]
\centering
\caption{tpAUC(50\%, 5\%) scores by pre-trained classifier and news truthfulness. Bold values indicate the higher-scoring category for a classifier.}
\begin{tabular}{@{}ccc@{}}
\toprule
Classifier    & Fake  & Real  \\ \midrule
AICD          & 0.00  & \textbf{0.12}  \\
Binoculars    & \textbf{95.25} & 87.96 \\
DAIGT         & 58.35 & \textbf{80.22} \\
Desklib       & \textbf{97.63 }& 90.41 \\
Fakespot      & 91.24 & \textbf{95.55} \\
MAGE          & \textbf{22.47} & 0.00  \\
Pangram       & \textbf{99.11} & 94.81 \\
SuperAnnotate & 61.10 & \textbf{72.81} \\
e5-LoRA       & 66.18 &\textbf{ 71.91} \\ \midrule
Mean          & 65.70 & 65.98 \\ \bottomrule
\end{tabular}
\label{tab:news-pretrained-real}
\end{table}

For the DACTYL-trained classifiers, DeBERTa-V3-large (DXO) once again had the highest tpAUC(80\%, 5\%) scores of 99.85 and 99.99 for both real and fake news, respectively, suggesting a negligible difference between the two categories. 

\begin{figure}[h!]
    \centering
    \includegraphics[width=1\linewidth]{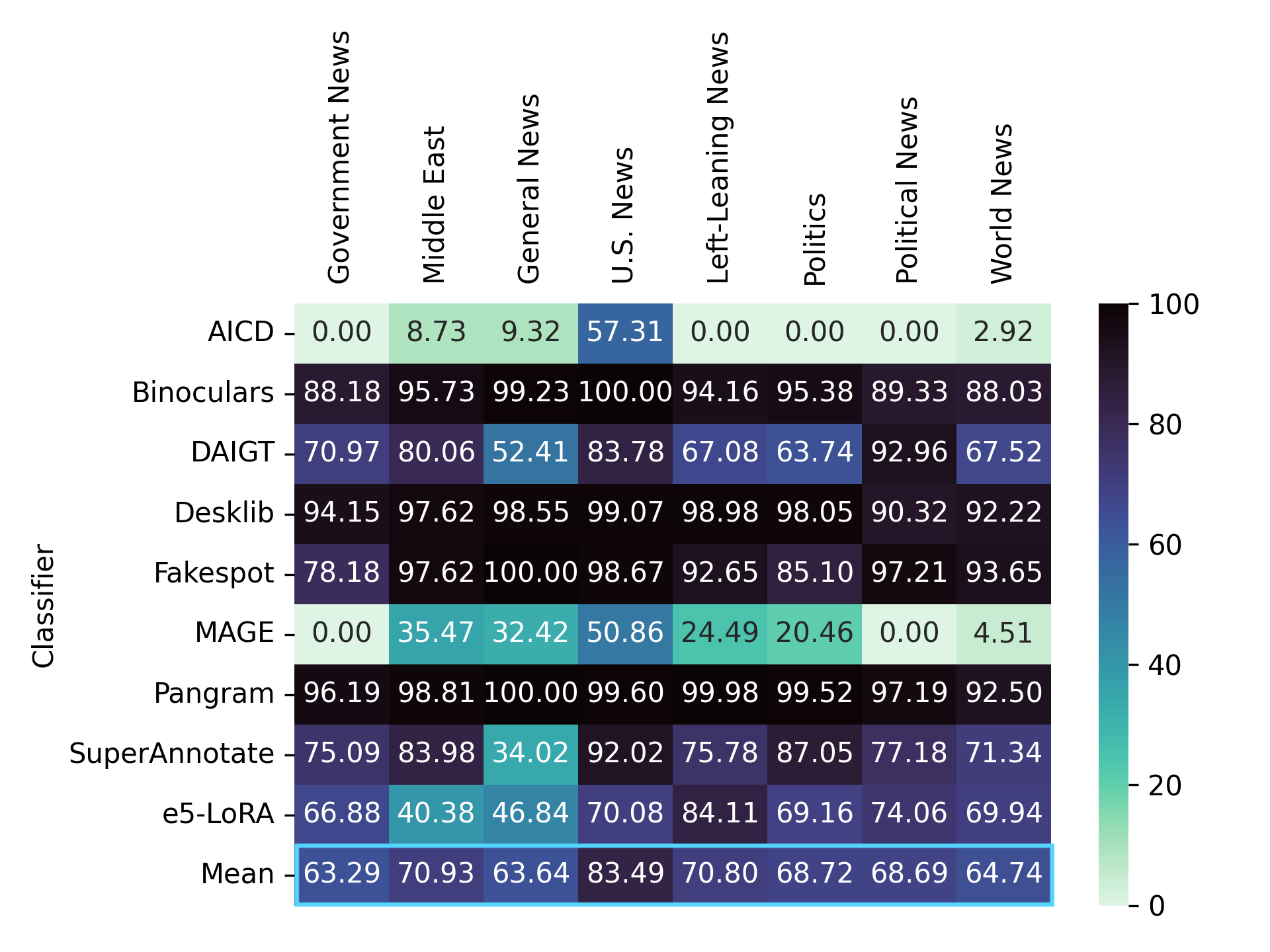}
    \caption{tpAUC(50\%, 5\%) scores by news topic for pre-trained classifiers.}
    \label{fig:news-pretrained-topic}
\end{figure}

Stratifying by news topics exhibited similar behavior (Figure \ref{fig:news-pretrained-topic}). However, the US news category was noticeably easier to detect, with a tpAUC score nearly 13 points higher than the next easiest topic (Middle East news). 

The lack of difficulty for the US news topic might be more indicative of training set bias rather than an inherent issue for specific topics. DACTYL-trained classifiers achieved perfect tpAUC(80\%, 5\%) scores in all topics except for Politics (99.96 by DeBERTa-V3-large DXO and BCE) and Political News (98.86 by ModernBERT-base DXO and BCE and distilRoBERTa-base DXO and BCE). 

The country of the news outlet style in the instruction prompt appears to influence detection slightly. We exclude MAGE and AICD as they achieved tpAUC scores of 0 across all news outlets. The pre-trained classifiers' mean tpAUC(50\%, 5\%) scores for the eight news outlets ranged from 81 to 86. The UK news outlets have a lower mean score, ranging from 81.65 to 82.90, while the US news outlets have a slightly higher range, from 83.89 to 85.5. This gap is likely due to the UK news style using British spelling conventions, an attack referred to as an alternative spelling attack in the RAID dataset \citep{dugan_raid_2024}. The Desklib and Fakespot classifiers, trained on RAID, do not suffer from performance drops when switching between US and UK news outlets. However, SuperAnnotate and e5-LoRA, trained on subsets of RAID, show apparent differences between the two groups. 
\begin{figure}[h!]
    \centering
    \includegraphics[width=1\linewidth]{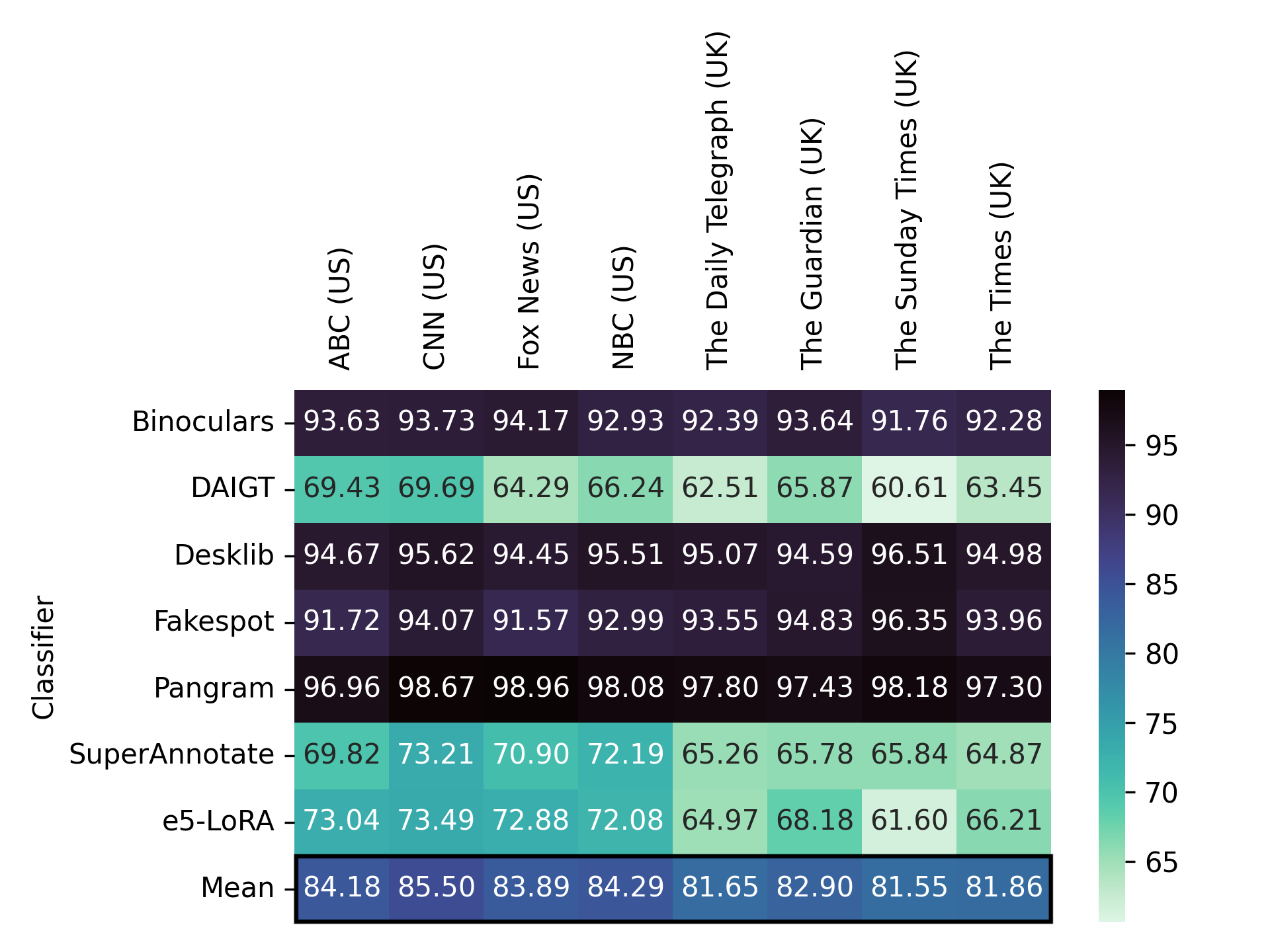}
    \caption{tpAUC(50\%, 5\%) scores by news outlet style for the pre-trained classifiers.}
    \label{fig:news-pretrained-newsoutlet}
\end{figure}

We did not observe a similar effect for the DACTYL-trained classifiers. DeBERTa-V3-large (DXO) obtained the ``lowest'' maximum tpAUC(80\%,5\%) of any of the eight news outlets (99.93 for The Daily Telegraph UK). All other news outlets had at least one classifier achieve a perfect tpAUC score. 

\begin{figure}[h!]
    \centering
    \includegraphics[width=1\linewidth]{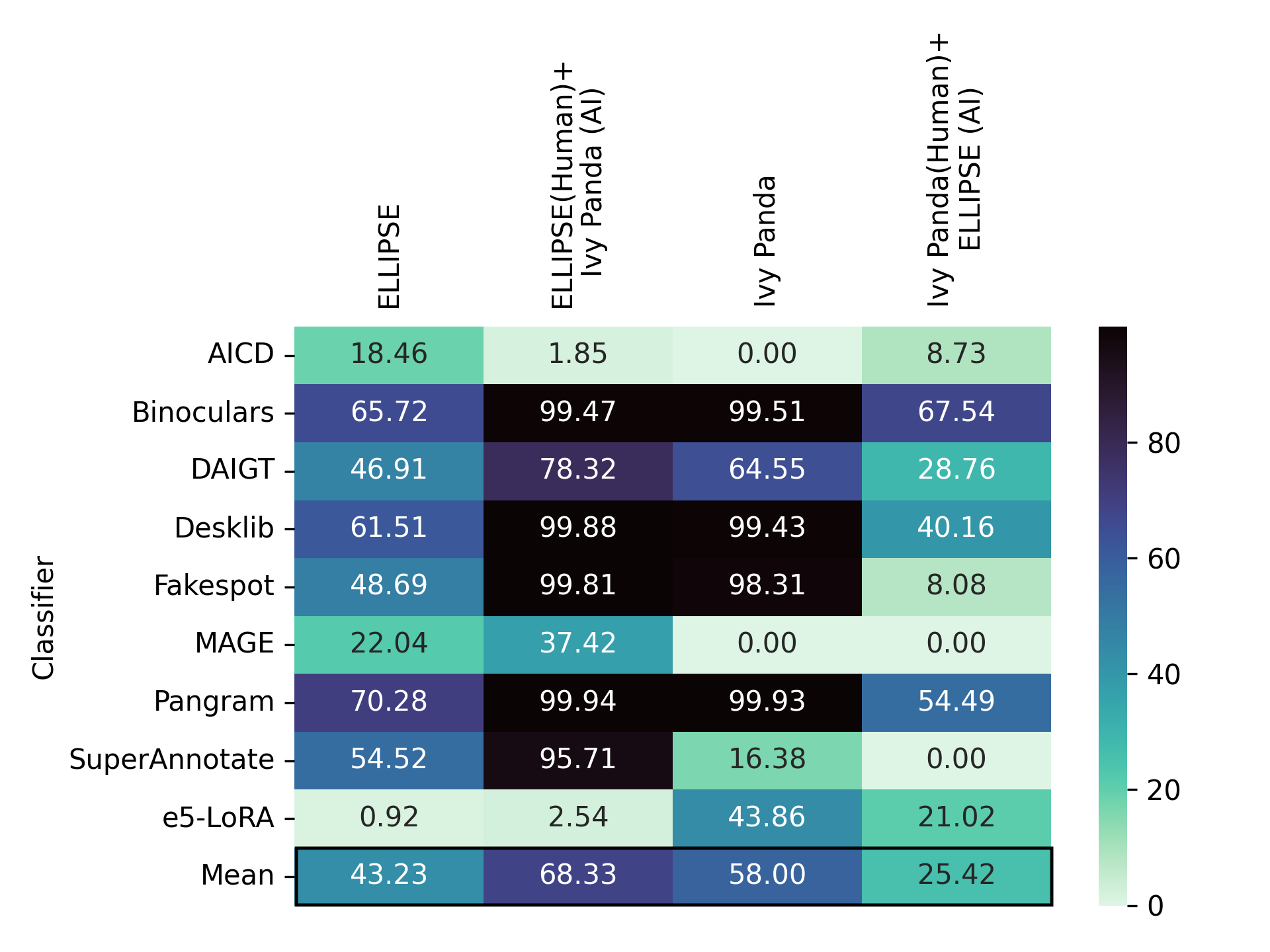}
    \caption{tpAUC(50\%, 5\%) scores for student essays for the pre-trained classifiers.}
    \label{fig:essays-pretrained-classifier}
\end{figure}

\subsubsubsection{Student Essays}
We evaluate the two subsets of student essays, Ivy Panda and ELLIPSE. ELLIPSE has two categories of interest: human essays written by English language learners and AIG essays generated in the style of those learners. We investigate if classifiers are biased toward this ``ELL class.'' We evaluate four possible combinations of Ivy Panda and ELLIPSE texts (results in \ref{fig:essays-pretrained-classifier}):

\begin{itemize}
    \item ELLIPSE Human + ELLIPSE AI
    \item ELLIPSE Human + Ivy Panda AI
    \item Ivy Panda Human + Ivy Panda AI
    \item Ivy Panda Human + ELLIPSE AI
\end{itemize}

ELLIPSE AIG essays were noticeably difficult: the mean tpAUC scores were below 50. Ivy Panda AIG essays are easy to distinguish, with a mean tpAUC of at least 58. Multiple classifiers had a tpAUC score greater than 99 in test combinations involving those AIG essays. 

Fortunately, this weakness can be trained against, based on Table \ref{tab:essays-trained-classifier}. 

\begin{table}[h!]
\centering
\caption{Highest tpAUC(80\%, 5\%) scores on the student essays test set.}
\begin{tabular}{@{}ccccccc@{}}
\toprule
Human Subset & AIG Subset & tpAUC & pAUC  & AUC   & AP    & Classifier              \\ \midrule
Ivy Panda    & ELLIPSE    & 98.19 & 99.81 & 99.97 & 99.83 & distilRoBERTa-base(DXO) \\
ELLIPSE      & ELLIPSE    & 98.25 & 99.82 & 99.96 & 99.93 & distilRoBERTa-base(DXO) \\
ELLIPSE      & Ivy Panda  & 99.95 & 99.99 & 100.0 & 100.0 & ModernBERT-base(DXO)    \\
ELLIPSE      & Ivy Panda  & 99.95 & 99.99 & 100.0 & 100.0 & ModernBERT-base(BCE)    \\
Ivy Panda    & Ivy Panda  & 99.98 & 100.0 & 100.0 & 100.0 & ModernBERT-base(DXO)    \\
Ivy Panda    & Ivy Panda  & 99.98 & 100.0 & 100.0 & 100.0 & ModernBERT-base(BCE)    \\
Ivy Panda    & Ivy Panda  & 99.98 & 100.0 & 100.0 & 100.0 & DeBERTa-V3-large(BCE)   \\ \bottomrule
\label{tab:essays-trained-classifier}
\end{tabular}
\end{table}

The ELLIPSE AIG texts are still slightly difficult to detect, but the gap is relatively small, even at a higher TPR.

\subsection{Adversarial Results}

We explore whether our CPT (continued pre-trained) models can evade detection better than the base Llama 3.2 1B instruct model can for the pre-trained classifiers. Then, we evaluate DACTYL-trained classifiers by comparing the performance of DXO and BCE classifiers.  

We use the same human texts from the non-adversarial test set for evaluation.

\subsubsection{CPT vs Base Models}
We observed that the tpAUC(50\%, 5\%) scores were mostly zero for most pre-trained classifiers; thus, we relaxed $\alpha$ and $\beta$ to 40\% and 30\%, respectively. However, there were still some zero tpAUC scores (SuperAnnotate did not have a non-zero tpAUC in any domain/model combination), so we use the ranking system mentioned in section \ref{sec:eval-classifier-performance}: rank first by tpAUC, followed by pAUC, AUC, and finally AP score. For pAUC, we also use a maximum FPR of 30\% for a consistent $\beta$ value. We determine if the CPT model was harder to detect than the base model in a particular domain if it maintained a lower tpAUC score (followed by tie-breakers if needed in the other three metrics). For example, in Table \ref{tab:daigt-cpt-vs-base-results}, the CPT model for reviews has a much lower tpAUC (33.54) than the base model for reviews (44.5). Thus, we can claim the CPT model benefited from the continued pre-training, at least for the DAIGT classifier. In contrast, the CPT news model was easier to detect than the base model: 94.51 vs 72.54.

\begin{table}[h!]
\centering
\caption{DAIGT results on the adversarial test set.}
\begin{tabular}{@{}cccccc@{}}
\toprule
Model Type & Domain    & tpAUC(40\%, 30\%) & pAUC(30\%)  & AUC   & AP    \\ \midrule
CPT        & Reviews   & 33.54 & 72.33 & 83.64 & 4.87  \\
CPT        & Tweets    & 36.3  & 76.76 & 81.53 & 43.87 \\
Base       & Tweets    & 43.37 & 79.32 & 84.13 & 46.77 \\
Base       & Reviews   & 44.5  & 77.14 & 87.07 & 6.63  \\
CPT        & Abstracts & 63.82 & 86.68 & 92.59 & 58.85 \\
Base       & Essays    & 67.93 & 88.37 & 92.26 & 45.62 \\
CPT        & Essays    & 70.87 & 89.1  & 93.76 & 34.75 \\
Base       & News      & 72.54 & 89.65 & 92.73 & 54.24 \\
Base       & Abstracts & 76.42 & 91.63 & 95.06 & 77.12 \\
Base       & CW        & 83.84 & 94.09 & 96.59 & 47.43 \\
CPT        & CW        & 94.09 & 97.64 & 98.79 & 50.37 \\
CPT        & News      & 94.51 & 97.6  & 98.78 & 72.35 \\ \bottomrule
\end{tabular}
\label{tab:daigt-cpt-vs-base-results}
\end{table}

We indicate which CPT models across which domains posed a greater challenge to pre-trained classifiers in Table \ref{tab:cpt-vs-base}. CPT models excelled at the three shorter text domains (tweets, reviews, and abstracts), degrading performance compared to the base model for all nine classifiers. CPT models outperformed the base models for six classifiers in the news and essays domains. However, the CPT model for creative writing struggled against the base model, with only three classifiers performing worse on the CPT model's generations.

\begin{table}[h!]
\centering
\caption{CPT vs base model results on pre-trained classifiers. \cmark\ indicates that the CPT model was harder to detect; \xmark\ indicates that the base model was harder.}
\begin{tabular}{@{}ccccccc@{}}
\toprule
Classifier    & Tweets & Reviews & Abstracts & News   & Essays & CW    \\ \midrule
AICD          & \cmark& \cmark& \cmark& \cmark& \xmark& \xmark\\
Binoculars    & \cmark& \cmark& \cmark& \cmark& \cmark& \xmark\\
Desklib       & \cmark& \cmark& \cmark& \xmark& \cmark& \xmark\\
e5-LoRA       & \cmark& \cmark& \cmark& \cmark& \cmark& \cmark\\
DAIGT         & \cmark& \cmark& \cmark& \xmark& \xmark& \xmark\\
Fakespot      & \cmark& \cmark& \cmark& \cmark& \cmark& \xmark\\
MAGE          & \cmark& \cmark& \cmark& \cmark& \xmark& \cmark\\
Pangram       & \cmark& \cmark& \cmark& \xmark& \cmark& \xmark\\
SuperAnnotate & \cmark& \cmark& \cmark& \cmark& \cmark& \cmark\\ \bottomrule
\end{tabular}
\label{tab:cpt-vs-base}
\end{table}

\subsubsection{Defending Against CPT Models}
\label{sec:defending-against-cpt-models}
\begin{figure}[h!]
    \centering
    \includegraphics[width=1\linewidth]{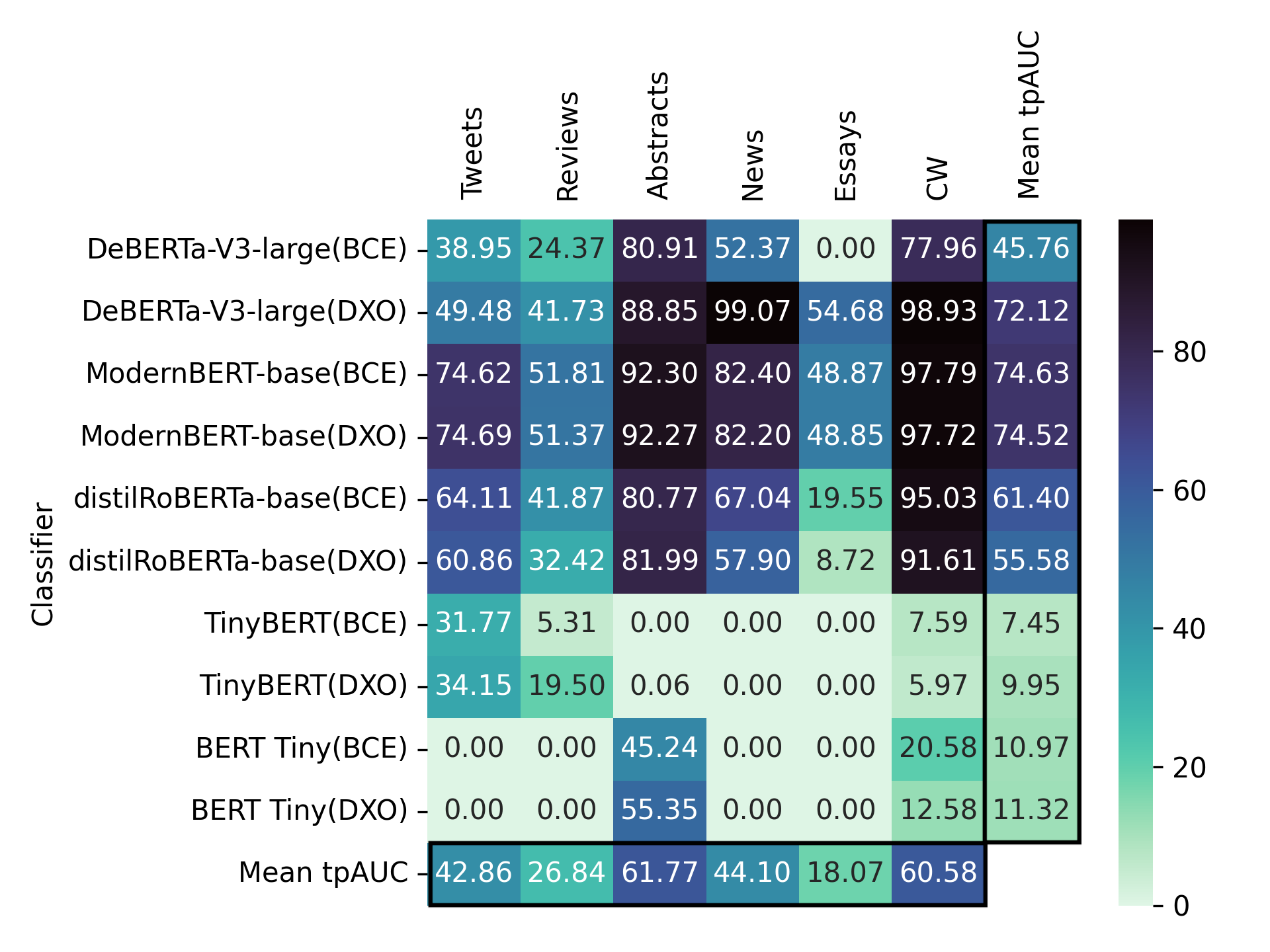}
    \caption{tpAUC(50\%, 5\%) scores by classifier and domain on the adversarial (CPT) test set.}
    \label{fig:cpt-dacty-trained-by-domain}
\end{figure}

Our DACTYL training set includes CPT model-generated texts, although these were not the same models used in testing. We compare mean tpAUC(50\%, 5\%) scores across domain and classifier type (BCE vs DXO) in Figure \ref{fig:cpt-dacty-trained-by-domain}, similar to Figure \ref{fig:dacty-trained-non-adv}. We exclude the base Llama 3.2 1B Instruct texts.

Note that the highest mean tpAUC score is just under 75, a relatively high gap to the 90+ tpAUC scores we saw for the non-adversarial results. However, some classifiers performed strongly on some domains: DeBERTa-V3-large (DXO) achieved a 99.07 tpAUC score on the CPT news model, suggesting that classifiers have the \textit{potential} to defend against an unseen fine-tuned/CPT model. 

DXO classifiers had mixed results compared to their BCE peers. For the DeBERTa-V3-large, TinyBERT, and BERT-tiny, using the tpAUC loss function improved their mean tpAUC scores (although for the smaller classifiers, the gains were more modest). ModernBERT and distilRoBERTa saw a decrease in their scores after training with the tpAUC loss function. The DXO classifiers that scored less on the non-adversarial test set showed improvements on the CPT test set. Conversely, the DXO classifiers that showed improvements on the non-adversarial test set struggled more on the CPT test set than their BCE counterparts. These results indicate a possible trade-off between performance on adversarial and non-adversarial test sets. This phenomenon has been well documented in the literature, referred to as the robustness-accuracy trade-off \citep{tsipras2018robustness}. However, this phenomenon does not apply to DeBERTa-V3-large (DXO) --- it is the only classifier to show improvements on both non-adversarial and adversarial tests over the BCE-trained version. We suspect the discrepancy might be due to the larger model size and having more parameters improves generalization. 

\subsection{Evaluating on the OOD Test Set}

Similar to the pre-trained classifiers' struggles on the DACTYL test set, we investigate if our DACTYL-trained classifiers struggle on the OOD test set using tpAUC(40\%, 30\%) scores, as shown in Table \ref{tab:ood-test-results}. For reference, we include the DACTYL test set (excluding the generations from the (non-CPT) Llama 3.2 1B Instruct). For all sub-test sets, we compute the scores on that entire subset (i.e., no averaging by domain).  

\begin{table}[h!]
\centering
\caption{tpAUC(40\%, 30\%) scores on the OOD test set. DAA refers to the DREsS and AIG-ASAP dataset. Bold values indicate the highest scores.}
\begin{tabular}{@{}cccccccc@{}}
\toprule
Classifier & type & AuText. & BEEMO & DACTYL & DAA & MAGE & Mean tpAUC \\ \midrule
DeBERTa-V3-large   & DXO  & 0.92            & \textbf{25.72} & \textbf{98.56}  & \textbf{88.89  }        & 5.64 & \textbf{43.95 }     \\
DeBERTa-V3-large   & BCE  & 0.0             & 23.82 & 96.17  & 59.6           & 0.95 & 36.11      \\
ModernBERT-base    & DXO  & 1.81            & 16.46 & 98.4   & 49.57          & 7.13 & 34.67      \\
ModernBERT-base    & BCE  & 1.78            & 22.8  & 98.4   & 27.25          & \textbf{7.15} & 31.48      \\
distilRoBERTa-base & BCE  & 1.54            & 4.03  & 98.01  & 39.37          & 7.01 & 29.99      \\
distilRoBERTa-base & DXO  & 1.1             & 1.8   & 97.36  & 30.75          & 4.39 & 27.08      \\
TinyBERT           & DXO  & \textbf{2.38}            & 14.8  & 91.94  & 19.17          & 0.0  & 25.66      \\
TinyBERT           & BCE  & 0.67            & 10.77 & 92.22  & 16.37          & 0.0  & 24.01      \\
BERT Tiny          & DXO  & 0.42            & 16.88 & 91.8   & 6.33           & 0.17 & 23.12      \\
BERT Tiny          & BCE  & 0.0             & 11.3  & 91.99  & 5.21           & 0.0  & 21.7       \\ \bottomrule
\end{tabular}
\label{tab:ood-test-results}
\end{table}

There are three main observations: (1) as classifier size grows, the mean tpAUC score increases; (2) DXO classifiers have slightly better generalizability compared to their BCE peers (except for distilRoBERTa-base); (3) a classifier trained on one domain may not generalize to other texts within the same domain (see the low scores on the AuTexTification subset). 

The first observation highlights a vital scaling property --- ideally, larger classifiers should generalize better \citep{urbizu-etal-2023-scaling}. This trend suggests that as a model's parameters increase, they should learn better representations of the data (i.e., they do not just memorize patterns). That is the case for the DACTYL-trained classifiers, at least based on the mean tpAUC score. However, individual subsets, such as AuTexTification and MAGE, do not appear to follow this scaling law. This gap can be due to DACTYL containing only one-shot/few-shot text generation and mixcase examples, which might not translate well to zero-shot generations. The increased performance on DREsS and AIG-ASAP might be due to similarities between ELLIPSE and Ivy Panda, as ELLIPSE and DREsS contain ELL texts. 

The second observation indicates the general advantage of DXO. The results support the claim that optimizing tpAUC directly, rather than using binary cross-entropy as a proxy, can lead to better performance and even generalization. For the lone exception, distilRoBERTa-base, we suspect that our training parameters might not be effective; we might be able to obtain a higher tpAUC score with some tuning, as \cite{yuan2023libauc}'s experiments involved hyperparameter tuning. 

The last observation seems to be an example of the robustness-accuracy trade-off we saw in section \ref{sec:defending-against-cpt-models}. The AuTexTification dataset contains no adversarial attacks or challenging texts --- texts are generated via a continuation prompt similar to some of MAGE's texts (although MAGE includes some paraphrased texts as adversarial texts)\citep{sarvazyan2023overview} \citep{li_mage_2024}. In contrast, the AIG-ASAP dataset contains adversarially modified AIG texts. Yet, to a certain extent, most classifiers do better on the DREsS/AIG-ASAP subset (and even BEEMO) than the former two. We attribute this performance difference to the adversarial nature of BEEMO and AIG-ASAP. 

\subsection{Deployment Scenario --- Student Essays}

While the four X-risks (tpAUC, pAUC, AUC, and AP) often give a general idea of classifier performance, these metrics summarize performance over a range of thresholds. When classifiers are deployed in real-world settings, decision-makers often want to see a classifier's performance given a fixed threshold. We simulate a deployment of two DACTYL-trained classifiers for detecting AIG student essays --- ModernBERT-base BCE (MB-BCE) and DeBERTa-V3-large DXO (DV3L-DXO). For our ``real-world'' test set, we use the DAA subset (DREsS/AIG-ASAP). This selection for our test set offers a more pessimistic view of our classifiers.  

We selected MB-BCE as it had the highest tpAUC(50\%,5\%) among all trained classifiers on the non-adversarial student essays domain. DV3L-DXO was the best DXO classifier overall on the adversarial DACTYL test set for student essays. 

\subsubsection{Threshold Selection}

We evaluate two possible cases for threshold selection using either the ROC or PRC curves. We rescale our thresholds from 0 to 1 to 0 to 100, as some determined thresholds were small. We list how we decide each threshold for each case. For consistency, a positive sample is an AIG essay, and a negative sample is a human-written essay.

Regardless of which curve is selected, most educators are interested in minimizing false positives --- students wrongly accused of using AI \citep{turnitinUnderstandingFalse}. Therefore, we focus on selecting desirable levels for each curve using FPR (ROC) or precision (PRC).  FPR and precision rely on minimizing false positives, but they approach them differently. FPR evaluates the number of false positives relative to the total number of negative samples. Precision measures the number of false positives relative to all samples \textit{flagged} as positive.

We determine our thresholds using the DACTYL student essay test set. Making a threshold based on the ``test set'' is reasonable, as we have access to our labels (as most AIG text developers should). However, we cannot make thresholds using the DAA test set, as it is unlikely that both educators and developers would have access to real-world labels. 

\subsubsubsection{ROC}
Individual points on the ROC curve map to individual TPR@FPR values. As mentioned, the ROC curve contains TPR@FPR values with no viability in a real-world setting (such as TPR@FPR=100\%). To determine the threshold using the ROC curve, we select a maximum FPR $\beta$ and calculate the threshold with the FPR that is closest (but less than) to $\beta$ \citep{krishna_paraphrasing_2023}. 
\subsubsubsection{PRC}
For the PRC, we set a minimum precision $p_r$ and choose the threshold with the highest recall (TPR) $\geq p_r$. We use PyTorch's \verb|torcheval| library to calculate the threshold for a fixed precision\citep{paszke2019pytorch}. 

\subsubsection{Results by Threshold}
We report four calculated thresholds by classifier: TPR@FPR=5\% \citep{dugan_raid_2024}, TPR@FPR=1\%\citep{krishna_paraphrasing_2023}, recall@precision=95\%, and  recall@precision=99\% in Table \ref{tab:dactyl-student-essays-thresholds}. We include a fifth fixed threshold of 50 (0.5) as a baseline. Note that the MB-BCE classifier generally has higher scores across all metrics. The MB-BCE is confident in its predictions of human texts, judging by the low thresholds. The DV3L-DXO is more cautious in its threshold selection, with a relatively high threshold at 95.11 for recall@precision=99\%. 
\begin{table}[h!]
\small
\centering
\caption{DACTYL test set derived thresholds and metrics for student essays. Bold values indicate the best values for each classifier.}
\begin{tabular}{@{}ccccccc@{}}
\toprule
Classifier    & Method                                & Threshold &  FPR & TPR/Recall & Precision & macro-F1 \\ \midrule
\multirow{5}{*}{MB-BCE}   & Threshold=50.0                        & 50        & \textbf{0.09 }    & 94.33           & \textbf{99.75}          & 97.94         \\
                          & TPR@FPR=5.0\%                         & 0.02      & 4.86     & \textbf{98.75}           & 88.27          & 95.25         \\
                          & Recall@Precision$\geq$95.0\% & 0.26      & 1.89     & 97.85           & 95.04          & 97.54         \\
                          & TPR@FPR=1.0\%                         & 1.19      & 0.99     & 96.91           & 97.3           & 98.02         \\
                          & Recall@Precision$\geq$99.0\% & 7.3       & 0.36     & 95.9            & 99             & \textbf{98.25  }       \\ \midrule
\multirow{5}{*}{DV3L-DXO} & Threshold=50.0                        & 50        & 1.77     & 97.16           & 95.32          & 97.41         \\
                          & TPR@FPR=5.0\%                         & 9.85      & 4.97     & \textbf{98.58 }          & 88.02          & 95.09         \\
                          & Recall@Precision$\geq$95.0\% & 45.96     & 1.89     & 97.38           & 95.03          & 97.38         \\
                          & TPR@FPR=1.0\%                         & 70.48     & 0.99     & 96.22           & 97.29          & 97.78         \\
                          & Recall@Precision$\geq$99.0\% & 95.11     & \textbf{0.35}     & 94.61           & \textbf{99.01}& \textbf{97.8}\\ \bottomrule
\end{tabular}
\label{tab:dactyl-student-essays-thresholds}
\end{table}

Unfortunately, the MB-BCE classifier's low threshold fails to generalize with extremely high FPR values greater than 90 on the DAA dataset, as seen in Table \ref{tab:daa-results-thresholds}. 
\begin{table}[h!]
\small
\centering
\caption{DACTYL-determined thresholds on the DAA dataset.}
\begin{tabular}{@{}ccccccc@{}}
\toprule
Classifier                & Method                                & Threshold & FPR   & TPR/Recall & Precision & macro-F1  \\ \midrule
\multirow{5}{*}{MB-BCE}   & Threshold=50.0                        & 50        & \textbf{91.41}& 85.72      & 54.81     & 40.2           \\
                          & TPR@FPR=5.0\%                         & 0.02      & 94.69 & \textbf{94.12}& \textbf{56.25}& 39.91          \\
                          & Recall@Precision$\geq$95.0\% & 0.26      & 93.38 & 91.74      & 55.96     & 40.4           \\
                          & TPR@FPR=1.0\%                         & 1.19      & 92.88 & 90.14      & 55.66     & \textbf{40.36}\\
                          & Recall@Precision$\geq$99.0\% & 7.3       & 92.37 & 88.22      & 55.27     & 40.19          \\ \midrule
\multirow{5}{*}{DV3L-DXO} & Threshold=50.0                        & 50        & 5.91  & 91.43      & 95.24     & \textbf{92.5}\\
                          & TPR@FPR=5.0\%                         & 9.85      & 15.61 & \textbf{94.59}& 88.68     & 89.86          \\
                          & Recall@Precision$\geq$95.0\% & 45.96     & 6.62  & 91.74      & 94.72     & 92.36          \\
                          & TPR@FPR=1.0\%                         & 70.48     & 2.98  & 88.98      & 97.48     & 92.44          \\
                          & Recall@Precision$\geq$99.0\% & 95.11     & \textbf{0.66}& 84.16      & \textbf{99.4}& 90.76          \\ \bottomrule
\end{tabular}
\label{tab:daa-results-thresholds}
\end{table}

The DV3L-DXO's cautious thresholds appear to benefit it, as its highest FPR is 15.61\%. However, this FPR is still high in deployment: for example, Vanderbilt University submitted over 75,000 (human) essays to Turnitin in 2022 \citep{vanderbiltGuidanceDetection}. Ultimately, if we used this threshold (TPR@FPR=5\% for DV3L-DXO), we could expect around 11,708 false positives. Using a lower acceptable FPR of 1\%  reduces the DAA FPR to just under 3\%. Setting a higher precision of 99\% also works: the FPR is down to 0.66\% and the TPR is at a respectable 84.16\%. 

The threshold selection process also showcases the trade-off between catching more AIG texts and avoiding false positives. A higher threshold naturally reduces FPR, but at the expense of TPR --- you are more likely to miss actual AIG texts. We emphasize that threshold selection is highly domain-dependent. In our deployment scenario, we are interested in minimizing false positives. Thus, we are more inclined to pick a higher threshold. However, for Amazon reviews, the cost of letting an AIG review through (such as a fake one-star review) could damage a seller's reputation and a consumer's trust. Likewise, a developer training an LLM may accept a high FPR to avoid training an LLM on AIG texts. 

This deployment scenario also underscores the dangers of overfitting to the test set and not properly evaluating generalizability. One might naively opt to deploy the MB-BCE classifier solely because it achieves the highest tpAUC score on the essays domain. As we can see, the classifier overfits significantly on DACTYL and may be learning superficial features between AIG and human texts. The DXO classifier does not showcase this behavior.

\section{Conclusion}
\subsection{Limitations}
While our analysis of our datasets and detectors intends to be exhaustive, we mention some limitations. First, our dataset only includes one-shot/few-shot examples (excluding CPT continuation generations) but does not include more traditional zero-shot generated texts. However, we did so as most literature includes mostly zero-shot generations, so we addressed this gap via our dataset. Given the classifiers' lack of generalizability to the MAGE and AuTexTification datasets, we suspect that learning on few-shot generations may not translate directly to zero-shot generations. Our dataset only focuses on English texts, but we caution that detecting AIG texts in other languages might yield novel problems. 

We also point out that we selected our training parameters based on prior experiments; it is likely that with further hyperparameter tuning, some of the DXO classifiers can outperform their BCE counterparts \citep{yuan2023libauc}. For our simulated deployment, we did not quantitatively factor in the costs of false positives and negatives for simplicity. In some real-world settings, we would have access to such details (e.g., time/money spent on falsely accusing a student of using an LLM), which could affect the threshold identification process. Also, we did not include additional factors that could boost detection; for example, fake news articles often spread faster than real news articles \citep{mitStudyTwitter}. Including metrics such as the number of shares in a fixed timeframe could help better discern between human and AIG articles. 

Finally, AIG text detectors are imperfect, and decision-makers (such as educators) should carefully weigh context before deciding whether a text is AIG. While the end goal is to build a robust detector that does not need a ``human-in-the-loop'', the current detectors are not fully ready. 

\subsection{Summary}
We provide an intensive analysis of building an AIG text detector. We highlight inconsistencies in AIG text evaluation and focus on the X-risk binary classification measures: tpAUC, pAUC, AUC, and AP. We then construct the large-scale one-shot/few-shot DACTYL dataset covering six highly vulnerable domains. We also include texts from full-parameter CPT Llama 3.2 1B Instruct models. Existing AIG text detectors struggle on our dataset, suggesting that generalizability is not guaranteed. CPT-generated texts also evade detectors better than non-CPT generations of the same model.  We also observe several vulnerabilities in several domains: AIG one-star reviews are harder to detect, AIG news articles written in the style of UK news outlets lead to some degradation, and AIG essays written in the style of ELL students also evade detection. Fortunately, we demonstrate that DACTYL-trained classifiers can close the gap in these vulnerabilities. We point out that CPT generations still seem challenging to detect, but DACTYL-trained classifiers perform much better than the pre-trained ones. However, we note that classifiers trained with binary cross-entropy seem to overfit to the test set, while DXO classifiers tend to generalize to unseen texts better. We confirm this trend by evaluating a mock deployment scenario using the DAA dataset between the binary cross-entropy trained ModernBERT-base and the tpAUC-optimized DeBERTa-V3-large classifiers.

While AIG text detection seems daunting, with newer LLMs and attacks released monthly, we observe some evidence of generalizability in existing classifiers. For example, multiple classifiers had little trouble with DeepSeek-V3 despite being a more recent LLM. We also want to highlight Pangram, the only pre-trained classifier to achieve a tpAUC(50\%, 5\%) greater than 70 on the DACTYL non-adversarial test set. Pangram's success indicates that it is possible to build a robust classifier.

\subsection{Future Work}
DACTYL is certainly not meant to be a finalized dataset --- AIG text detection and evasion continue to evolve rapidly. This development speed necessitates continuous updates to AIG text datasets to avoid severe distributional shifts between training and real-world texts. Future work for ``DACTYL 2.0'' might explore full-parameter instruction tuning, rather than just continued pre-training. We also recommend exploring LLM fine-tuning as multiple LLM providers have convenient (but costly) fine-tuning APIs.

The six domains are just a few of several impacted domains. AIG code is now a new concern --- a Georgetown University study demonstrated that several LLMs' generated code contains security vulnerabilities \citep{georgetownCybersecurityRisks}. Workflows such as GitHub pull requests, a process of merging code from different branches in a repository, could use AIG text detection to flag LLM-generated code for further inspection. Wikipedia is also facing concerns of AIG articles, with an estimated 5\% (in August 2024)  of English articles generated by LLMs \citep{brooks-etal-2024-rise}. \cite{brooks-etal-2024-rise} argue that these articles are typically low-quality. 

Besides including more data, modifications to the classifier could show improvements in robustness. Kolmogorov-Arnold Networks (KANs) have demonstrated improvements over traditional multi-layer perceptrons (MLPs) \citep{liu2024kan}. Exploring advancements in architecture (such as ModernBERT) is essential to improving classification performance. 

As with many security frontiers in technology, AIG text detection is an arms race involving multiple defenders from academia, governments, industry, and various individual and organized attackers. Many traditional machine-learning techniques that saw success in text classification tasks may not translate into AIG text detection. We encourage future researchers (and detector developers) to thoroughly inspect their machine-learning processes and classifiers. We expect adversaries to do the same. By discovering their own classifiers' vulnerabilities and limitations before launching them, defenders can stay ahead in the race.

\section*{Acknowledgments}

This work was performed using resources provided by the Cambridge Service for Data Driven Discovery (CSD3) operated by the University of Cambridge Research Computing Service (\url{www.csd3.cam.ac.uk}), provided by Dell EMC and Intel using Tier-2 funding from the Engineering and Physical Sciences Research Council (capital grant EP/T022159/1), and DiRAC funding from the Science and Technology Facilities Council (\url{www.dirac.ac.uk}). 

The Department of Computer Science and Technology provided credits for GPU resources for CSD3's Wilkes3 GPU cluster. Pangram Labs provided access to their commercial AI text detector. Haneul Yoo granted access for the DReSS dataset to evaluate classifier performance. 

The second author is supported by Cambridge University Press \& Assessment.

\microtypesetup{protrusion=false}
\bibliographystyle{unsrt}
\bibliography{references}  






\end{document}